\documentclass{scrartcl}
\usepackage[utf8]{inputenc}
\usepackage[most]{tcolorbox}
\usepackage{setspace}

\usepackage[backend=biber, 
style=authoryear, 
maxcitenames=1,
maxbibnames=3,
uniquename=init, 
uniquelist=false,
natbib=true]
{biblatex}
\DeclareDelimFormat[bib]{finalnamedelim}{\addspace\&\space}

\DeclareDelimFormat{nameyeardelim}{\addcomma\space}
\DeclareDelimFormat[bib]{nameyeardelim}{\addspace}
\DeclareNameAlias{sortname}{family-given}
\DeclareFieldFormat[article, inbook, incollection, inproceedings, misc, thesis, unpublished]{title}{#1\isdot}
\renewbibmacro{in:}{}

\let\oldcite\cite
\renewcommand{\cite}[1]{(\oldcite{#1})}

\appto{\bibsetup}{\singlespacing}

\usepackage{pdflscape}
\usepackage{booktabs}
\usepackage{multirow}
\usepackage{placeins}

\usepackage[bookmarks=false]{hyperref}
\urldef{\huggingurl}\url{https://tinyurl.com/5n8bjnbh}

\usepackage{svg}
\usepackage{enumitem}
\usepackage{authblk}

\usepackage{tikz}
\usepackage{pgfplots}
\usepgfplotslibrary{groupplots}
\usepackage{subcaption}
\usetikzlibrary{matrix}

\usepackage[T1]{fontenc}
\usetikzlibrary{fit, positioning}

\usetikzlibrary{shapes,arrows}
\usetikzlibrary{positioning, shapes.geometric, calc}

\pgfplotsset{compat=1.18}

\usepackage{hhline,booktabs}

\usepackage [most] {tcolorbox}
\usepackage{xcolor}

\definecolor{mydarkblue}{RGB}{0, 0, 139}
\definecolor{myroyalblue}{RGB}{65, 105, 225}
\definecolor{mydodgerblue}{RGB}{30, 144, 255}
\definecolor{mynavyblue}{RGB}{0, 0, 128}
\definecolor{myskyblue}{RGB}{135, 206, 235}
\definecolor{mycornflowerblue}{RGB}{100, 149, 237}

\newtcolorbox[]{myfigure}[1][]{
          colback=gray!20, colframe=gray!80}

\providecommand{\keywords}[1]
{
  \small	
  \textbf{Keywords:} #1
}

\title{Clinical information extraction for Low-resource languages with Few-shot learning using Pre-trained language models and Prompting}

\author[a,b,c,d,e]{Phillip Richter-Pechanski\thanks{Corresponding author: PRP, phillip.richter-pechanski@med.uni-heidelberg.de.}}
\author[a,c,d,e]{Philipp Wiesenbach}
\author[c]{Dominic M. Schwab}
\author[c]{Christina Kiriakou}
\author[c,e]{Nicolas Geis}
\author[a,c,d,e]{Christoph Dieterich\thanks{AF and CD jointly supervised this work and are shared last authors.}}
\author[b]{Anette Frank\protect\footnotemark[2]}

\affil[a]{Klaus Tschira Institute for Integrative Computational Cardiology, Heidelberg, Germany}
\affil[b]{Department of Computational Linguistics, Heidelberg University, Germany}
\affil[c]{Department of Internal Medicine III, University Hospital Heidelberg, Germany}
\affil[d]{German Center for Cardiovascular Research - Partner site Heidelberg/Mannheim, Germany}
\affil[e]{Informatics for Life, Heidelberg, Germany}

\date{}

\begin{document}

\maketitle

\newpage

\begin{abstract}
    \noindent\textbf{Abstract} 
    
    \noindent A vast amount of clinical data is still stored in unstructured text. Automatic extraction of medical information from these data poses several challenges: high costs of required clinical expertise, restricted computational resources, strict privacy regulations, and limited interpretability of model predictions.
    
    \noindent Recent domain adaptation and prompting methods using lightweight masked language models showed promising results with minimal training data and allow for application of well-established interpretability methods. We are first to present a systematic evaluation of advanced domain adaptation and prompting methods in a low-resource medical domain task, performing multi-class section classification on German doctor’s letters. We evaluate a variety of models, model sizes, (further-pre)training and task settings, and conduct extensive class-wise evaluations supported by Shapley values to validate the quality of small-scale training data, and to ensure interpretability of model predictions.
    
    \noindent We show that in few-shot learning scenarios, a lightweight, domain-adapted pretrained language model, prompted with just 20 shots per section class, outperforms a traditional classification model, by increasing accuracy from $48.6\%$ to $79.1\%$. By using Shapley values for model selection and training data optimization we could further increase accuracy up to $84.3\%$. Our analyses reveal that pretraining of masked language models on general-language data is important to support successful domain-transfer to medical language, so that further-pretraining of general-language models on domain-specific documents can outperform models pretrained on domain-specific data only.

    \noindent Our evaluations show that applying prompting based on general-language pre-trained masked language models combined with further-pretraining on medical-domain data achieves significant improvements in accuracy beyond traditional models with minimal training data. Further performance improvements and interpretability of results can be achieved, using interpretability methods such as Shapley values. Our findings highlight the feasibility of deploying powerful machine learning methods in clinical settings and can serve as a process-oriented guideline for clinical information extraction projects working with low-resource languages.

\end{abstract} 

\keywords{Prompting, Few-shot learning, Pretraining, Medical information extraction, Doctor's letters, Language models}

\section{Introduction}
    \label{intro}
    Vast amounts of clinical data are stored in unstructured text, such as doctor's letters. Natural language processing (NLP) and machine learning (ML) 
    can make their information available for research and clinical routine. While  supervised ML approaches rely on large amounts of manually annotated training data, recent developments in NLP showed promising results in text classification tasks using pretrained language models (PLM) and prompting \citep{brown2020language}. Prompting exploits the ability of PLMs to make correct predictions if guided through context; in combination with supervised methods they achieve state-of-the-art results on various text classification tasks \citep{liu2023pre}.
    
    Doctor’s letters are typically divided into sections, such as anamnesis, diagnosis or medication, containing semantically related sentences.
    Typically, it is not necessary to consider all sections to obtain specific medical information, \citep{richter2021automatic} or medication information \cite{uzuner2010extracting}. Instead, medical information extraction (MIE) tasks, such as medication extraction or patient cohort retrieval, can be improved by contextualizing the information in a doctor's letter \cite{edinger2017evaluation}. However, automatic section classification is non-trivial due to a high variability of the structuring of information across physicians and time periods \citep{lohr2018cda}.
    
\begin{figure}
        \centering
        \includegraphics[width=\textwidth]{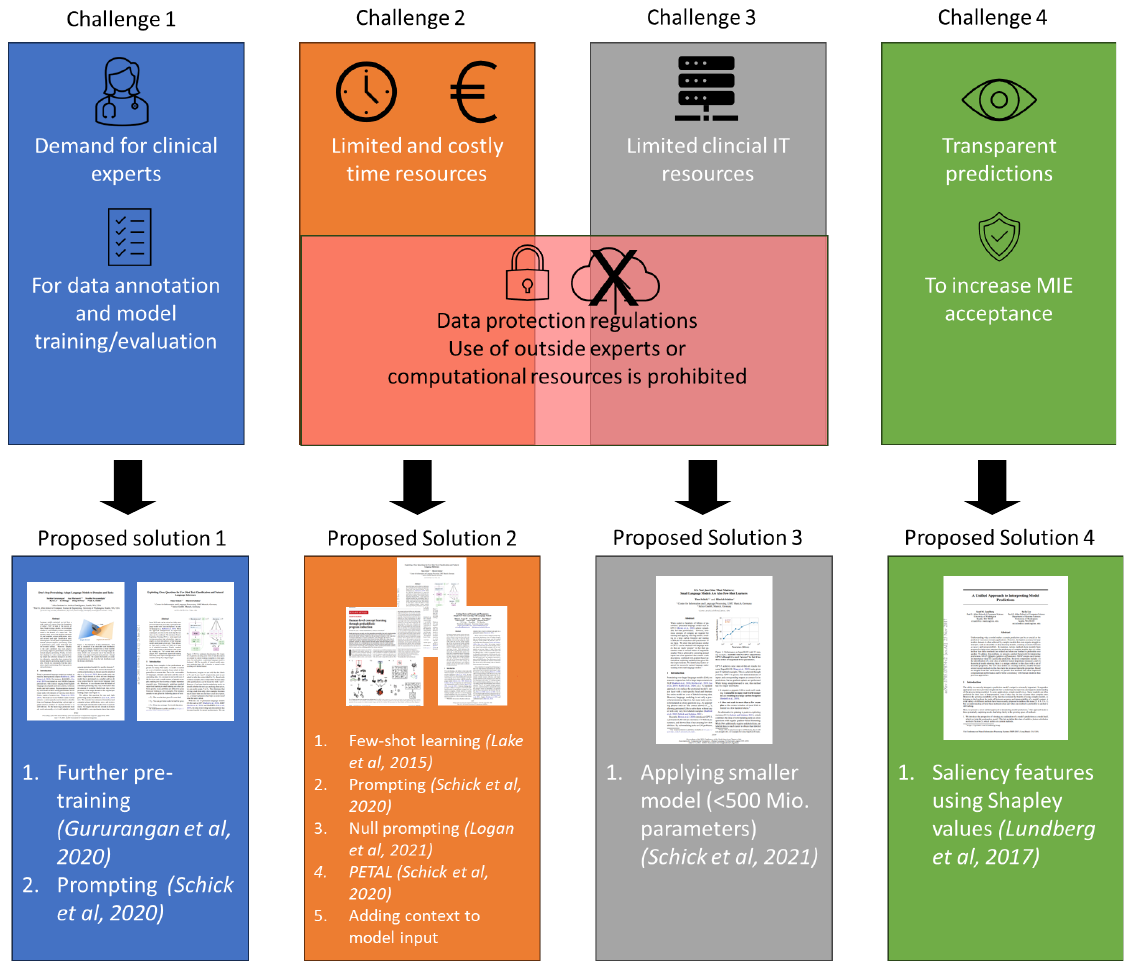}
        \caption{\textbf{Challenges for MIE projects in clinics}: Our proposed solutions on main challenges for MIE projects in a clinical setting.}
        \label{fig:intro_motivation}
    \end{figure}

    In close collaboration with physicians from clinical routine, we identified four challenges of MIE projects in the clinical domain \citep{hahn2020medical} (Fig. \ref{fig:intro_motivation}).
    \begin{enumerate}[label=Ch$_\arabic*$]
        \item \textbf{Domain-and-Expert-dependent}: Annotation projects often require an active involvement of clinical domain experts for data annotation and model evaluation. This is particularly relevant for low-resource languages such as German.
        \item \textbf{Resource-constrained}: Domain experts are costly and have only limited time resources. By contrast, external expert involvement is difficult due to strict data protection measures\cite{richter2021automatic}.  
        \item \textbf{On-premise}: Personal data are confidential, which means that many MIE projects are carried out entirely on premise i.e., in the clinical IT infrastructure. However, computational resources in clinical infrastructures are often a limiting factor \cite{taylor2023clinical}.
        \item \textbf{Transparency}: Due to the sensitivity of clinical information, safety standards for using MIE results in clinical routine are high: model predictions must be of high quality, transparent, explainable and as comprehensible as possible \cite{tjoa2020survey}.
    \end{enumerate}
    We evaluate best-practice strategies to identify an ideal setup to address the multifaceted challenges of conducting a MIE task such as clinical section classification. Specifically, we identify and propose the following solutions:
    \begin{enumerate}[label=S$_\arabic*$]
        \item We reduce the demand for clinical knowledge in MIE by exploiting existing domain knowledge available in hospitals, such as clinical routine documents. We evaluate domain- and task-adapted  \citep{gururangan2020don} general-use PLMs, as well as PLMs pretrained on clinical data from scratch \citep{bressem2023medbert} in combination with prompt-based learning methods \citep{Schick2020pet}, which require only limited training data.
        \item To reduce time investment and costs of manual data annotation through clinical experts, we apply few-shot learning \citep{lake2015human} and context-enriched training data using prompt-based fine-tuning with Pattern-Exploiting Training (PET) \citep{Schick2020pet} and compare the results with supervised sequence classification methods. We further evaluate the feasibility of null prompts \citep{logan2021cutting}, which have been shown to alleviate the search for effective prompts while achieving improved results.
        \item While large language models (LLMs) have recently shown impressive medical capabilities \cite{singhal2023large}, their demands of compute power, and currently unsolved issues regarding automatic evaluation, faithfulness control and trustworthiness make their use in clinical contexts often impractical \cite{parnami2022learning,thirunavukarasu2023large}. 
        We therefore focus on smaller PLMs ($110$-$345$ million learnable parameters) in a few-shot learning setting. Notably, prompt-based fine-tuning already achieves higher accuracy with smaller, encoder-based  PLMs compared to PLMs fine-tuned for sequence labeling with full-fledged training dataset \cite{Schick2020pet}.
        \item To address the need for transparent and trustworthy model predictions in clinical routine, we use well-established masked-language-models. They allow application of state-of-the-art interpretability methods that rely on saliency features computed with, e.g., Shapley values \citep{Lundberg_NIPS2017_7062}, to explain our model predictions.
    \end{enumerate}
    In what follows we conduct in-depth evaluations of these proposed solutions in a real-world section classification task, applied to German doctor’s letters from the cardiovascular domain. To our knowledge, this is the first in-depth evaluation of a prompt-based fine-tuning method such as PET on real-world clinical routine data in German language.

    % Figures and tables

    \subsection{State of research}
        % Prompting, few shot
        \textbf{From fine-tuning to few-shot learning with prompting.} Since 2017, most NLP tasks apply a \textit{pretrain-then-finetune} paradigm: neural models are pretrained with a language modeling objective and then fine-tuned for a down-stream task. However, even fine-tuned PLMs often perform poorly with sparse training data \cite{gao-etal-2021-making} and require 
        significant amounts of manually labeled training data to perform well
        \cite{liu2023pre}. Especially with low(er)-resource languages and in special domains, we often face a scarcity of high-quality labeled data.  
        With recent scaled-up language models, we observe another shift to a \textit{pretrain-then-prompt} paradigm, where tasks are formulated using natural language prompts \citep{Schick2020pet,reynolds2021prompt,gao-etal-2021-making,shin2020autoprompt}, revealing impressive zero-shot capabilities of these models \cite{liu2023pre,kojima2205large}. 
        While in many applications at least a few training samples are still required to guide model predictions, prompt-based learning soon matched and even surpassed the performance of fine-tuning in various few-shot learning settings \citep{liu2023pre,taylor2023clinical}.
        
        Although model size played a critical role in this development \cite{chowdhery2022palm}, smaller, encoder-based PLMs have also been successfully applied in few-shot scenarios using prompt-based fine-tuning in combination with a semi-supervised approach \citep{schick-schutze-2021-just,wang2022towards}. Especially, framing text classification tasks as cloze-style problems using Pattern-Exploiting Training (PET) showed promising results for various classification tasks \cite{schick-schutze-2022-true}.
        
        % Further pre-training
        \textbf{Domain adaptation through \textit{further-pretraining}.} Pretrained LMs achieve high performance across many tasks \cite{sun2019fine}, yet it typically drops in out-of domain settings. Several studies explored \textit{further-pretraining} on domain-specific data \cite{zhu-etal-2021-pre} in such cases, demonstrating that further-pretraining even on small-sized task-specific data can improve results in out-of-domain down-stream tasks \citep{gururangan2020don}.
        
        % Medical PLMs
        \textbf{PLMs for medical domain.} Medical PLMs, pretrained on medical data from scratch and further-pretrained medical PLMs have outperformed general PLMs in several tasks \citep{sivarajkumar2022healthprompt,taylor2023clinical}. However, clinical routine texts, as used in this study, have unique textual properties compared to biomedical texts on which such models are trained. This increases the complexity of medical NLP tasks in clinical routine \citep{leaman2015challenges,hahn2020medical}. Also, only a limited number of further-pretrained and clinical PLMs have been published to date, mostly for English, primarily due to strict data protection regulations \cite{lee2020biobert,li2023comparative,bressem2023medbert}.
          
        % Clinical prompting
        \textbf{Prompting methods in clinical NLP.} Despite extensive research on PLMs for medical domain, previous research has mainly focused on supervised fine-tuning with full-fledged training data approaches that use large amounts of training data \cite{wu2020deep, taylor2023clinical} with the exception of \citep{taylor2023clinical} who investigated prompting on English clinical data. Thus, there is a need to investigate how further-pretraining influences prompting methods in few-shot scenarios.
        
        PET performed well in various downstream tasks in English, i.a.\ 
        in biomedical text classification, where for adverse drug effect classification it outperformed GPT3 with an $F1$-score of $82.2\%$ vs.\ $68.6\%$ \citep{schick-schutze-2022-true}. This highlights the need for thorough evaluating PET in clinical routine tasks with medical domain PLMs, particularly for low-resource languages such as German \citep{leaman2015challenges,hahn2020medical}.
        
        % Section classification
        \textbf{Clinical section classification.} Identifying sections in clinical texts has been shown to enhance performance on several MIE tasks \citep{pomares2019current}. However, most studies focus on English clinical texts \cite{denny2008development,edinger2017evaluation}, and in-depth studies of prompting in few-shot learning scenarios are still lacking. Also, the limited availability of German PLMs \cite{bressem2023medbert} for clinical domains requires investigation of 
        suitable (further-)pretraining methods and their impact on prompting.
        
        % Interpretability
        \textbf{Interpretability.} 
        Given the black-box nature of deep learning architectures, the interpretability of model outputs is challenging and attracts much interest \cite{fan2021interpretability}, especially in safety-critical domains such as clinical routine. Various feature attribution methods have been developed to address these issues \cite{ribeiro2016should,sundararajan2017axiomatic,Lundberg_NIPS2017_7062},
        but we still face challenges in assessing their quality \cite{attanasio2022ferret}\cite{jacovi2020towards}. Shapley values provide a theoretically well-founded approach to determine the contribution of individual input features to a model prediction. A computationally optimized implementation called SHAP \cite{shapley1953value} can be applied out-of-the-box on transformer-based models. To our knowledge, we are the first to study the use of Shapley values for data and model optimization in clinical tasks. 
        
        % LLMs
        \textbf{Progress in the area of LLMs.} Recently, generative LLMs with billions of parameters deliver impressive results in various general \cite{brown2020language,chowdhery2022palm,workshop2023bloom,touvron2023llama} and biomedical and clinical NLP tasks \cite{singhal2023large,thirunavukarasu2023large,clusmann2023future,peng2023study}. However, many challenges need to be addressed before LLMs can be applied in clinical tasks \cite{wang2023large}: Running them via external APIs is typically prohibited due to data protection regulations. Despite efforts to make LLMs available for use in protected infrastructures (cf. \url{https://github.com/bentoml/OpenLLM}), model deployment in clinical infrastructures is often not feasible \cite{taylor2023clinical}. Moreover, out-of-the-box local GPT and Llama models have shown poor performance in biomedical tasks \cite{moradi2021gpt,wu2023pmc}. Finally, due to the generated outputs of autoregressive PLMs, their use in clinical NLP implies unsolved issues concerning automatic evaluation \cite{guo2023evaluating,chang2023survey} and judging the faithfulness of model predictions \cite{parcalabescu2023measuring}, which are both critical in the clinical domain.

        While evaluation of autoregressive LLMs will mature in the future, our study on encoder-based models %still 
        serves as a process-oriented guideline for MIE projects in clinical routine tasks for low-resource languages. All constraints discussed in this study: (1) expert-dependency, (2) data protection regulations, (3) demand for on-premise solutions and (4) transparency requirements, invariably apply to popular local LLMs such as Llama \cite{touvron2023llama} or Mistral \cite{jiang2023mistral}, and can serve as guidelines for evaluating these models, too.

\section{Methods}
    \label{section:methods}
    \subsection{Pattern-Exploiting Training (S\texorpdfstring{$_1$ \& S$_2$}{1 and 2})}

    In our experiments we systematically evaluate methods for few-shot learning, i.e., using minimal training data, in a clinical routine setting in a low-resource language, in our case German. Specifically, we evaluate PET, a semi-supervised prompting method optimized for few-shot learning scenarios \citep{Schick2020pet} which is designed to recast classical text classification or information extraction tasks as a language modeling problem. In our study we classify paragraphs of German doctor's letters into a set of nine section categories (Tab. \ref{tab:methods:count_section_classes_reduced}). The objective is, for instance, to accurately categorize a paragraph such as \textit{The patient reports pressure pain in the left chest} under the section class \textit{Anamnese}.

    To conduct PET experiments we need a pre-trained masked language model $M$ with a vocabulary $V$, a few-shot dataset with training instances $x_i \in X$ and target labels $y_i \in Y$. We further need a pattern function $P$ that maps instances to a set of cloze sentences (templates) $P: X \mapsto V^\ast$, and a verbalizer function $v: Y \mapsto V$ that maps each label to a single token from the vocabulary of $M$.
            
\begin{figure}
                \centering
                \includegraphics[width=\textwidth]{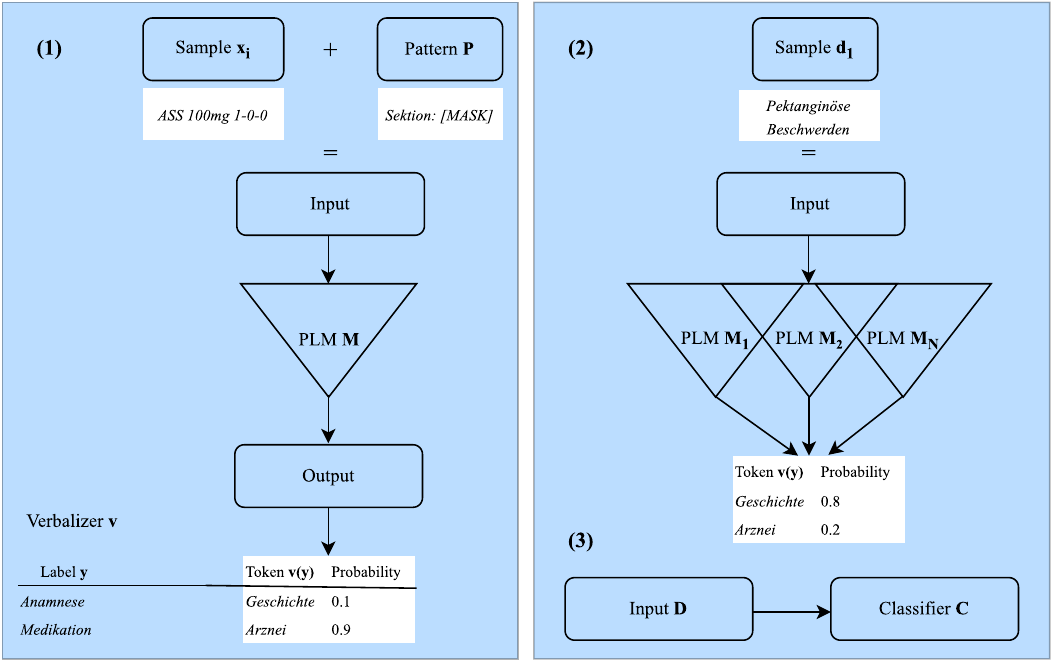}
                \caption{\textbf{PET workflow}: Three main steps: (1) Apply pattern function \textit{P(x)} to all few-shot training instances \textit{X}. Fine-tune a PLM \textit{M} using a language model objective on each pattern. The output of the PLM is mapped using a verbalizer function \textit{v(y)}. (2) An ensemble of \textit{M} trained on each pattern is used to annotate an unlabeled dataset \textit{D} with soft labels. (3) A classifier \textit{C} with a classification head is trained on \textit{D}.}
                \label{fig:methods:workflow}
    \end{figure}

    The PET workflow contains three basic steps (see Fig. \ref{fig:methods:workflow}): (1) applying $P$ to each input instance $x_i$ and fine-tune a model $M$ for each template to obtain the most likely token for the $MASK$ token $v(y)$, (2) use the ensemble of fine-tuned models $M$ from the previous step and annotate a large unlabeled dataset $D$ with soft labels and (3) train a final classifier C with a traditional sequence classification head on the labeled dataset $D$.

        \subsubsection{Creating templates}
            \label{section:methods:templates}
            Template engineering is a crucial hyperparameter in a PET experiment. For the \textit{core experiments} we used four different template types (examples in brackets):
            \begin{itemize}
                \item Null prompt: \texttt{SAMPLE [MASK]}\\ (\textit{Keine peripheren Ödeme} \texttt{[MASK]})
                \item Punctuation: \texttt{SAMPLE : [MASK]} and \texttt{SAMPLE - [MASK]}\\ (\textit{Keine peripheren Ödeme :} \texttt{[MASK]})
                \item Prompt: \texttt{SAMPLE Sektion [MASK]}\\ (\textit{Keine peripheren Ödeme Sektion} \texttt{[MASK]})
                \item Q\&A: \texttt{SAMPLE Frage: Zu welcher Sektion gehört dieser Text?\\ Antwort: [MASK]} \\(\textit{Keine peripheren Ödeme Frage: Zu welcher Sektion gehört dieser Text? Antwort:} \texttt{[MASK]})
            \end{itemize}
            To minimize engineering costs we also evaluated the feasibility of using exclusively null prompts, by removing all tokens from prompt templates, as proposed by \citep{logan2021cutting}. We defined three null prompt templates: (1) \texttt{SAMPLE [MASK]}; (2) \texttt{[MASK] SAMPLE} and (3) \texttt{[MASK] SAMPLE [MASK]}.
        \subsubsection{Verbalizer}
            Defining the verbalizer token can be tedious, because domain knowledge and technical expertise about the used PLM is required. This can be a significant issue, as such a comprehensive knowledge is uncommon in the clinical setting. Moreover, PET restricts the verbalizer token to a single token. Hence, an appropriate and intuitive token may not be applicable for a label mapping, if it is not included in the PLM's vocabulary. For instance, the word \textit{Anamnese} is not part of the \textit{gbert} vocabulary. This makes a verbalizer search for clinicians quite challenging. Therefore, we use PET with Automatic Labels (PETAL) for all our experiments, but the zero-shot baselines \cite{Schick2020petal}. This can reduce engineering costs and makes our experimental setup more comparable and reproducible.
            As visualized in Suppl. Fig. \ref{fig:suppl:methods:petal} 
            PETAL calculates the most likely verbalizer token per label, given the few-shot training data for each pattern and given a PLM. We created a separate verbalizer for each few-shot size for each training set.
            
    \subsection{Pretrained language models (S\texorpdfstring{$_1$ \& S$_3$}{1 and 3})}

        \label{section:methods:pre-training}
\begin{figure}
            \centering
            \includegraphics[width=\textwidth]{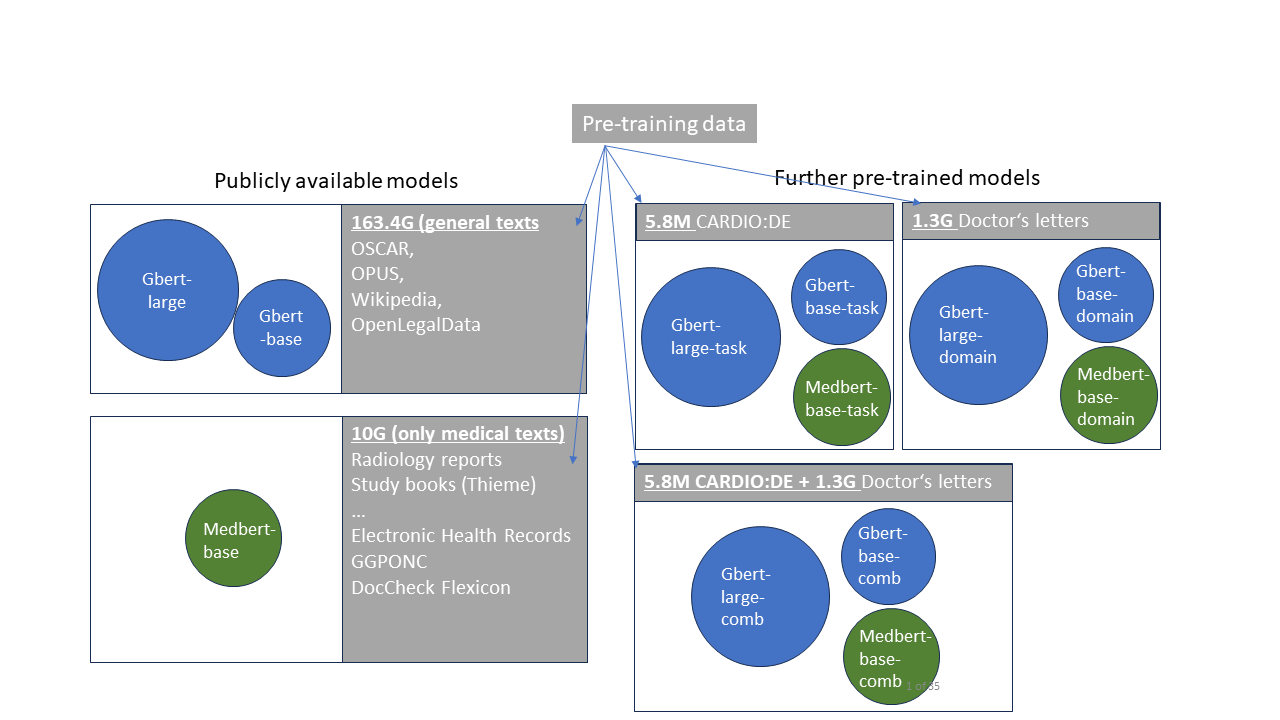}
            \caption{\textbf{Pretrained language models:} We use two publicly available PLMs: \textit{gbert} and \textit{medbertde}. We evaluate base and large \textit{gbert} models. Four pretraining methods are used: (1) publicly available, (2) task-adapted, (3) domain-adapted and (4) task- and domain-adapted combined.}
            \label{fig:methods:llm_pretraining}
        \end{figure}

        To evaluate the feasibility of exploiting existing clinical domain knowledge by 
        further-pretraining the used model before applying PET, we used a set of three language models, all based on the BERT architecture \citep{devlin2018bert} and available at Hugging Face Hub: (1) \texttt{deepset/gbert-base} \citep{chan2020german}, (2) \texttt{deepset/gbert-large} (\textit{gbert}), (3) \texttt{Smanjil/German-MedBERT} (\textit{medbertde}) \citep{bressem2023medbert}.
        For both \textit{gbert} and \textit{medbertde} we create medical-adapted variants by further pre-training, as proposed by \cite{gururangan2020don} to assess the impact of different pretraining datasets on section classification results (Fig. \ref{fig:methods:llm_pretraining}). We defined datasets for three different pretraining approaches:
        \begin{enumerate}
            \item \textit{task-adaptation.} Using CARDIO:DE cf. Section \ref{section:methods:annotated_data}. This data set contains unlabeled data extracted from the same source as the training and test data of the section classification task. It is relatively small, only $6M$. (PLMs appended with suffix \textit{-task})
            \item \textit{domain-adaptation.} Using 179,000 doctor's letters from Cardiology department at the University Hospital, cf. Section \ref{section:methods:pretraining_data}. This data set contains a broad range of texts from clinical routine in cardiovascular domain. With $1.3G$ it is significantly larger than the task-adaptation data set. (PLMs appended with suffix \textit{-domain})
            \item \textit{combination of both approaches} Further-pretrain a domain-adapted PLM on our task specific data (PLMs appended with suffix \textit{-comb})
        \end{enumerate}

        We performed pretraining using a masked language modelling objective (cf. \huggingurl). For hyperparameters see Suppl.\ Section \ref{section:suppl:hyperparameters}.  
    \subsection{Shapley values (S\texorpdfstring{$_4$}{4})}
        In many safety-critical domains, in particular in the clinical domain, it is crucial to (1) understand the inner workings of a model (faithfulness) and to (2) evaluate how convincing a model interpretation is for a human observer (plausibility) \cite{jacovi2020towards}. This can help to increase trust in model predictions (explainable AI) by identifying which token contributed to a specific prediction. Furthermore, if a model makes incorrect predictions, allocating such tokens can help to understand and address these issues.
        
        In recent years Shapley values became a valuable tool in NLP for local model interpretations using saliency features \cite{attanasio2022ferret}. Shapley values offer a systematic approach to attribute the impact of individual textual components (token, token sequences) on a model prediction. In our setup we apply Shapley values in two ways: (1) From a clinical routine perspective: to make deep learning model predictions more transparent and explainable, (2) From an engineering perspective: to detect biases or errors in the training data and to support choosing the most optimal model architecture. 
        Shapley values, originating from cooperative game theory, allocate the importance of each feature by averaging its marginal contribution across all possible feature combinations in predicting an outcome \citep{Lundberg_NIPS2017_7062}. The Shapley value for a feature \( i \) is given by:
        \begin{equation}
            \phi_i(f) = \sum_{S \subseteq N \setminus \{i\}} \frac{|S|!(|N|-|S|-1)!}{|N|!} \left[ f(S \cup \{i\}) - f(S) \right]
        \end{equation}
        Here \( f \) is the prediction function, \( S \) is a subset of all features without feature \( i \), and \( N \) is the set of all features.
        In our experiments we use SHAP (SHapley Additive exPlanations) because it offers an optimized algorithm that approximates Shapley values with reduced computational costs, making its application feasible for practical use \citep{mosca2022shap}.

    \subsection{Data}
        \subsubsection{Annotated corpus}
            \label{section:methods:annotated_data}
            For our experiments we used a German clinical corpus from the cardiovascular domain, \mbox{CARDIO:DE}, encompassing $500$ doctor's letters from the Cardiology Department at the Heidelberg University Hospital \cite{richter2023distributable, data/AFYQDY_2022}. The complete corpus contains $993,143$ tokens, with approximately $31,952$ unique tokens. The corpus was randomly split into \mbox{CARDIO:DE400} containing $400$ letters ($805,617$ tokens) for training and \mbox{CARDIO:DE100}, containing $100$ letters  ($187,526$ tokens) for testing. The corpus was automatically de-identified, by replacing protected health information (PHI) containing patient sensitive identifiers with placeholders using an in-house deep learning model \cite{richter2019deep}. This was followed by a manual review involving domain experts to fix de-identification errors. To increase readability and semantic consistency and to decrease the chance for re-identification, all PHI placeholders were replaced  with semantic surrogates, as proposed in \cite{lohr2021pseudonymization}.
            
            We split the corpus by newline characters, which are part of the \texttt{MS-DOC} source documents. Sentence splitting the corpus with publicly available sentence splitting methods or by pattern heuristics showed unsatisfactory results. Furthermore, sequence length of newline split paragraphs rarely exceed $512$ token  (min: $3$, max: $599$, mean: $30.9$, median: $16$, $99th$ percentile: $205$), thus, comply with most PLM sequence length restrictions.
            
            The corpus contains $116.898$ paragraphs manually annotated with $14$ section classes: \textit{Anrede, AktuellDiagnosen, Diagnosen, AllergienUnverträglichkeitenRisiken, Anamnese, AufnahmeMedikation, KUBefunde, Befunde, EchoBefunde, Labor, Zusammenfassung, Mix, EntlassMedikation, Abschluss} (see CARDIO:DE section classes, Suppl. Tab.\ \ref{tab:suppl:data:count_sect}). Manual annotation was conducted on paragraph level, no nested annotations were allowed. For our experiments we reduced the section classes to the most significant sections. We removed the \textit{Labor} section, as it contains flattened tables resulting in a large amount of relatively well structured and short numeric samples. Internal experiments showed that they can be sufficiently identified using regular expressions and patterns. Furthermore, we merged seven semantically similar classes in CARDIO:DE annotations to three meta classes: (1) \textit{Diagnosen}: (\textit{AktuellDiagnosen} $+$ \textit{Diagnosen}), (2) \textit{Medikation}: (\textit{AufnahmeMedikation} $+$ \textit{EntlassMedikation}) and (3) \textit{Befunde}: (\textit{KUBefunde} $+$ \textit{EchoBefunde} $+$ \textit{Befunde}). Our final dataset contains $49,258$ paragraphs annotated with $9$ section classes (Tab.\ \ref{tab:methods:count_section_classes_reduced}).

            \begin{table}
                \centering
                \begin{tabular}{lcc}
                    \toprule
                      & Training set & Test set \\
                    \midrule
                    Anrede & 402 & 99 \\
                    Diagnosen & 8,023 & 1,738 \\
                    AllergienUnverträglichkeitenRisiken & 1,031 & 236 \\
                    Anamnese & 1,188 & 281 \\
                    Medikation & 6,148 & 1,627 \\
                    Befunde & 15,396 & 3,914 \\
                    Zusammenfassung & 3,645 & 843 \\
                    Mix & 945 & 242 \\
                    Abschluss  & 2,805 & 695 \\
                    \midrule
                    Total & 39,583  & 9,675 \\
                    \bottomrule
                \end{tabular}
                \caption{\textbf{Distribution of section classes}: Number of samples per section class per corpus split.}
                \label{tab:methods:count_section_classes_reduced}
            \end{table}
            
            During annotation human annotators of CARDIO:DE were presented the whole document (for further annotation details, see \cite{richter2023distributable}). To mimic this setup for our automatic section classifiers in this study, we introduced basic information about document structure to the model without introducing additional pre-processing steps or external knowledge. In addition to our training data containing single paragraph samples we assessed two types of context-enriched datasets for our experiments (Examples Tab.\ \ref{tab:methods:data:context_types}):
            \begin{itemize}
                \item no-context (a single paragraph to be classified)
                \item context (previous paragraph + main paragraph + subsequent paragraph)
                \item prevcontext (previous paragraph + main paragraph)
            \end{itemize}

            \begin{table}
                \centering
                \begin{tabular}{lp{12cm}}
                    \toprule
                    Context type & Example \\
                    \midrule
                    nocontext & \textit{Cvrf: Hypertonie, Nikotinkonsum, Hypercholesterinämie} \\
                    context & \textit{- OP am 02.01.2011} \texttt{[SEP]} \textit{Cvrf: Hypertonie, Nikotinkonsum, Hypercholesterinämie} \texttt{[SEP]} \textit{Anamnese:} \\
                    prevcontext & \textit{- OP am 02.01.2011} \texttt{[SEP]} \textit{Cvrf: Hypertonie, Nikotinkonsum, Hypercholesterinämie} \\
                    \bottomrule
                \end{tabular}
                \caption{\textbf{Contextualized paragraphs:}  Examples of three different context types. We separate paragraphs using the \texttt{[SEP]} token.}
                \label{tab:methods:data:context_types}
            \end{table}
            
            The context-enriched samples still mostly comply with sequence length restrictions of PLMs (minimum $7$, maximum $967$, mean length $90.2$, median length $63$ and $99th$ percentile $371$ sub tokens).
        \subsubsection{Pretraining data} 
            \label{section:methods:pretraining_data}
            For all pretraining experiments we used an internal clinical routine corpus containing approximately $179,000$ German doctor's letters in a binary \texttt{MS-DOC} format covering the time period $2004$ to $2020$. We collected the letters from the Cardiology Department of the University Hospital Heidelberg. The pretraining corpus is disjoint from the annotated corpus. We conducted the following pre-processing steps: each letter was converted into a \texttt{UTF-8} encoded raw text file using the LibreOffice command line tool \texttt{soffice} (version 6.2.8). We chose LibreOffice, as it best preserved the structure of newlines and blanklines. We automatically de-identified all letters using a method based on a deep learning model trained on internal data, see \citep{richter2019deep}. We replaced PHI tokens with semantic surrogates, see \citep{lohr2021pseudonymization}. All doctor's letters were concatenated into a single raw text file. We separated each new letter by the sequence \texttt{\#\#\#BEGINN}. All empty lines and all tables containing laboratory values were removed. The corpus is sentence splitted using \texttt{NLTK}'s (version 3.7) \texttt{PunktSentenceTokenizer}.
            
            The doctor's letters were further supplemented by the GGPONC corpus, which contains German oncology guidelines, with a total of $2$ million token \citep{borchert2022ggponc}. The final corpus covers $1.3$ GB of raw text, approximately $218,084,190$ tokens and $667,903$ unique tokens.

    \subsection{Experimental setup}    
        \subsubsection{Metrics}
            \label{section:methods:metrics}            
            We measure section classification performance with accuracy for per-model results. In a multi-class text classification task, the accuracy is defined as the ratio of text documents correctly classified to their respective classes over the total number of text documents:
            \begin{equation}
                \text{Accuracy} = \frac{\sum_{i=1}^{n} \text{TP}_i}{\text{Total Number of Texts}}
            \end{equation}
            where \( \text{TP}_i \) represents the true predictions for each class \( i \) and \( n \) is the total number of classes.
            
            To measure section classification performance per-section class, we use the \( F_1 \)-score. It is defined as the harmonic mean of precision and recall given by
            \begin{equation}
                \text{Precision} = \frac{\text{TP}}{\text{TP} + \text{FP}}
                \end{equation}
                
                \begin{equation}
                \text{Recall} = \frac{\text{TP}}{\text{TP} + \text{FN}}
            \end{equation}
            Hence, the \( F_1 \)-score is defined by:
            \begin{equation}
                F_1 = 2 \times \frac{\text{Precision} \times \text{Recall}}{\text{Precision} + \text{Recall}}
            \end{equation}
            where TP, FP, and FN represent true positives, false positives, and false negatives, respectively.

            We used approximate randomization tests \citep{yeh2000more} to measure statistical significance for accuracy and $F1$-score results. Results are considered significant if $p<0.05$ \url{https://github.com/smartschat/art}.
        \subsubsection{Creating Few-Shot Data}
            \label{section:methods:few-shot-data}
            To conduct PET experiments we created six few-shot datasets. Each dataset contains N paragraphs per section class with size 
            N=10, 20, 50, 100, 200 and 400 randomly selected from the CARDIO:DE400 data (random seed $42$). Each paragraph includes the previous and subsequent context paragraph. All other context types (\textit{nocontext}, \textit{prevontext}) are derived from this dataset. Each few-shot set includes three labeled training files and three unlabeled files with the remaining samples from the CARDIO:DE400 dataset (Suppl. Fig. \ref{fig:suppl:experiments:fewshotfolder}). All experiments were evaluated on the complete CARDIO:DE100 held-out dataset.
        \subsubsection{Core Experiments}
            We conducted \textit{core experiments} to assess the performance of different section classification models along three dimensions to compare: (1)  fine-tuned sequence classification model variants (SC) to few-shot prompt-based learning with PET (S$_2$, Fig.\ \ref{fig:methods:workflow}), (2) four different pre-training methods for clinical adaptation (S$_1$), and (3) six different few-shot sizes: $10-400$ (S$_2$). 
            
            The SC model is trained using a BERT-architecture with an additional output layer for a sequence-classification task as described in  \cite{devlin2018bert}. We use the SC implementation of the PET framework, defined by the parameter \texttt{----method sequence\_classifier}.
            
            For all \textit{core experiments} we used base-sized BERT models (S$_3$) (\textit{gbert-base-*} and \textit{medbertde-base-*}) using all five templates combined and \textit{nocontext} samples (Suppl. Tab. \ref{tab:suppl:methods:core_experiments}). To measure standard deviation in \textit{core experiments} and \textit{additional experiments} we used three disjoint training sets including their unlabeled sets for each few-shot set. Furthermore we conducted all experiments with two random initital seeds ($123$ and $234$).
            
        \subsubsection{Additional experiments}
            \label{section:methods:additional_experiments}
            In \textit{additional experiments} we investigate the effectiveness of further parameters, using the model that performed best in \textit{core experiments}, with reduced few-shot sets: $20,50,100$ and $400$. We investigate the impact of (1) \textit{model size} comparing BERT-large and BERT-base models, (2) \textit{null prompt patterns}, and (3) \textit{contextualization}.
            In  \textit{core} and \textit{additional experiments} we further perform \textit{class-based evaluations} on two primary classes, which were selected with clinical experts: (1) \textit{Anamnese} (mostly unstructured) and (2) \textit{Medikation} (semi-structured).
            
            \textbf{Model size} (S$_3$): We evaluated the impact of adding model parameters, by comparing \textit{gbert-base} ($110$  million) vs \textit{gbert-large} ($340$ million) PLMs. We limited this setup to \textit{gbert} PLMs, since a large \textit{medbertde} was not published.
            
            \textbf{Null prompts} (S$_2$): \citep{logan2021cutting} discovered that the usage of null prompts, prompts without manually crafted templates achieve competitive accuracy to manually tuned prompt templates on a wide range of tasks. This is of particular interest in the clinical domain, to further reduce costly engineering efforts.
            
            \textbf{Adding context} (S$_2$): To introduce further information to the document structure, we added further context to each input sample to evaluate the effect of adding context paragraphs to each sample. We evaluated three types of context:
            \begin{enumerate}
                \item nocontext: \textit{Cvrf: Hypertonie, Nikotinkonsum, Hypercholesterinämie}
                \item context: \textit{- OP am 02.01.2011} \texttt{[SEP]} \textit{Cvrf: Hypertonie, Nikotinkonsum, Hypercholesterinämie} \texttt{[SEP]} \textit{Anamnese:}
                \item prevcontext: \textit{- OP am 02.01.2011} \texttt{[SEP]} \textit{Cvrf: Hypertonie, Nikotinkonsum, Hypercholesterinämie}
            \end{enumerate}

\section{Results}
    \subsection{Baselines}
\begin{figure}
        \centering
        \begin{tikzpicture}
        \begin{groupplot}[
            group style={
                group size=2 by 1,
                horizontal sep=10pt,
                ylabels at=edge left,
                group name=group,
            },
            legend style={
                anchor=north,
                yshift=-190pt,
                xshift=10pt,
                legend columns=-1,
                /tikz/every even column/.append style={column sep=0.5cm}
            },
            width=8cm, % Adjust the width as needed
            height=7cm,
        ]
    
    \nextgroupplot[
        ybar,
        enlarge x limits=0.48,
        ylabel={Accuracy},
        symbolic x coords={gbert,medbertde},
        xtick=data,
        nodes near coords,
        nodes near coords align={vertical},
        ymax=100,
        ymin=0,
    ]
    
        \addplot[fill=mydarkblue,nodes near coords style={rotate=90,left,color=black,yshift=0pt, xshift=28pt}] coordinates {(gbert,7.2) (medbertde,28.3)};
        \addplot[fill=myroyalblue,nodes near coords style={rotate=90,left,color=black,yshift=0pt, xshift=28pt}] coordinates {(gbert,13.0) (medbertde,12.8)};
        \addplot[fill=mydodgerblue,nodes near coords style={rotate=90,left,color=black,yshift=0pt, xshift=28pt}] coordinates {(gbert,7.2) (medbertde,15.4)};
        \addplot[fill=mynavyblue,nodes near coords style={rotate=90,left,color=black,yshift=0pt, xshift=28pt}] coordinates {(gbert,15.0) (medbertde,9.8)};
        \legend{public, task, domain, combined}
        
        \nextgroupplot[
            ybar,
            enlarge x limits=0.5,
            yticklabels={},
            symbolic x coords={gbert,medbertde},
            xtick=data,
            nodes near coords,
            nodes near coords align={vertical},
            ymax=100,
            ymin=0,
        ]    
        \addplot[fill=mydarkblue,nodes near coords style={rotate=90,left,color=white}] coordinates {(gbert,96.1) (medbertde,96.5)};
        \addplot[fill=myroyalblue,nodes near coords style={rotate=90,left,color=white}] coordinates {(gbert,96.6) (medbertde,96.5)};
        \addplot[fill=mydodgerblue,nodes near coords style={rotate=90,left,color=white}] coordinates {(gbert,96.5) (medbertde,96.6)};
        \addplot[fill=mynavyblue,nodes near coords style={rotate=90,left,color=white}] coordinates {(gbert,96.7) (medbertde,96.5)};
    
        \end{groupplot}
        \node[anchor=north] at (rel axis cs:0.7,-0.1) {(a)};
        \node[anchor=north] at (rel axis cs:1.75,-0.1) {(b)};
        \end{tikzpicture}

        \caption{\textbf{Section classification baseline results (lower/upper bound)}: We show accuracy scores per pre-training method (public, task-adapted, domain-adapted and combination of both) per model: \textit{gbert-base} and \textit{medbertde-base}. (a) Lower-bound: used in zero-shot prompting (b) Upper bound: \textit{full} training set.}
        \label{fig:results:baseline_results}
    \end{figure}
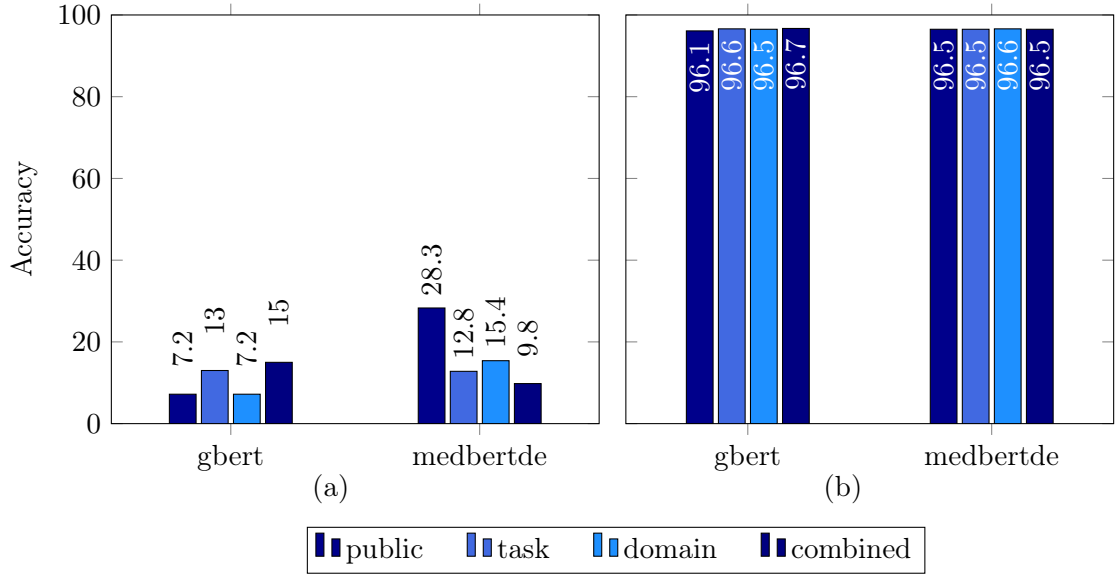

    We define two baselines to assess model performance in our \textit{core} and \textit{additional experiments}: as \textit{lower bound} we use a \textit{zero-shot prompting} approach; as  \textit{upper bound} we use a \textit{fine-tuned sequence classifier} trained on the \textit{full} size of the training corpus. 
    Fig.\ \ref{fig:results:baseline_results} shows the accuracy results for both baselines. The \textbf{upper bound} results exceed $96\%$ accuracy for both models. The further-pretrained \textit{gbert} models yield a minimal (statistically significant) advance of
    $0.4$-$0.6$ accuracy points above the original \textit{gbert-base}. For \textit{medbertde} no such difference is observed.
    
    The \textbf{zero-shot results} are all below $16\%$ accuracy, except for the public \textit{medbert-base}  that with $28.3\%$ achieves a great advance over \textit{gbert-base} with $7.2\%$ accuracy. However, the \textit{gbert} models further-pretrained on both task- and domain-specific data more than double the performance of the original model to $15\%$ accuracy, beyond \textit{gbert} pre-trained on domain-specific data only (\textit{*-domain}). All performance differences for \textit{gbert} are statistically significant, except \textit{gbert-base} and \textit{gbert-domain}.

    \subsection{Core experiments}  
\begin{figure}
        \centering
        \includegraphics[width=\textwidth]{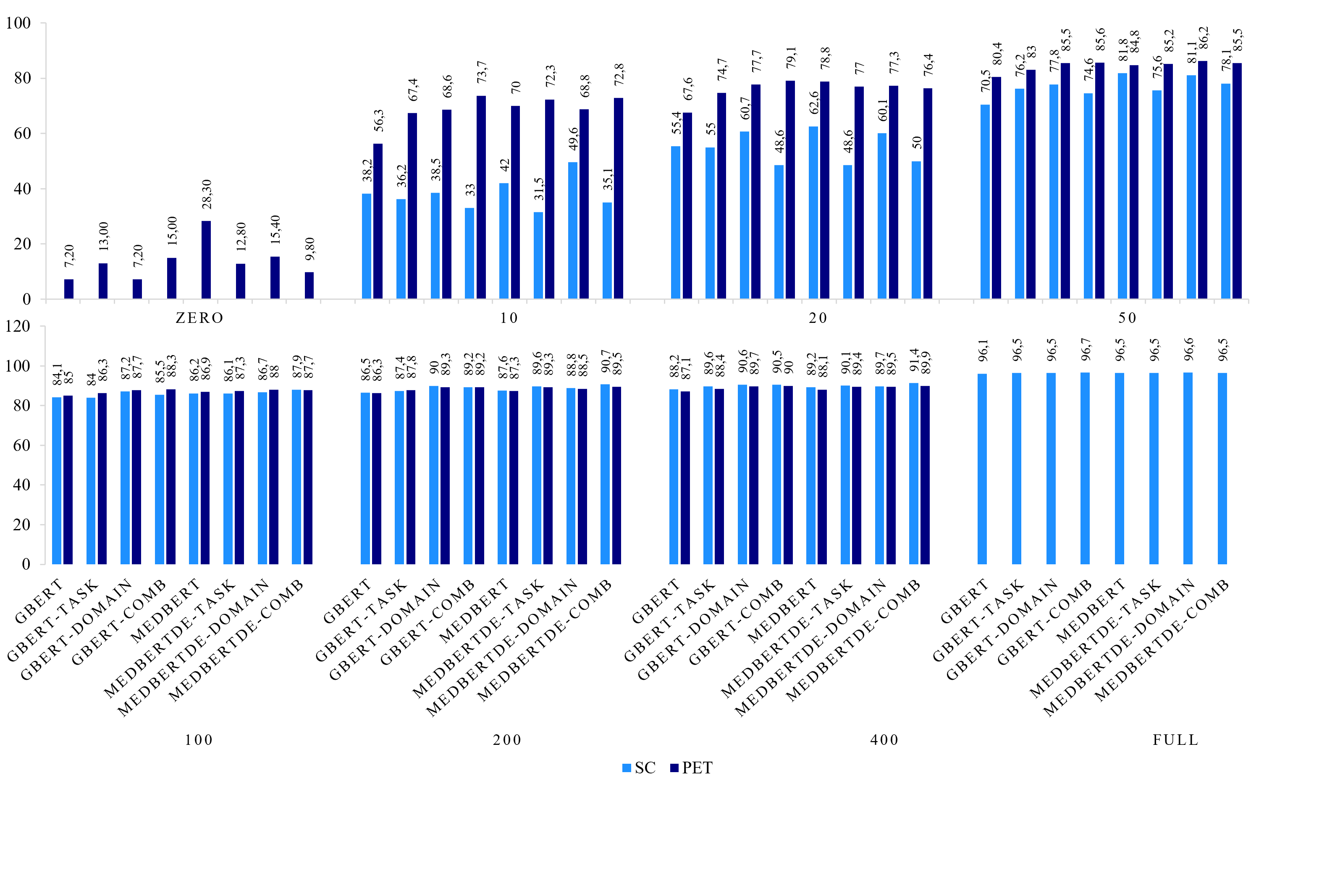}
        \caption{\textbf{Accuracy scores for core experiments and lower/upper bound}: comparing prompting using PET vs. SC, few-shot sizes $10-400$ and pre-training methods using base BERT models. For reference, lower-bound PET baselines trained with zero-shots (ZERO) and upper-bound SC models trained on complete training set (FULL).}
        \label{fig:results:core_experiments}
    \end{figure}

Fig.\ \ref{fig:results:core_experiments} presents our \textit{core experiment} results compared to the baselines introduced above.

        \textbf{PET vs.\ SC.}
        The PET model variants significantly outperform SC models at shot sizes $\le$100 in 31 out of 32 setups when comparing the same pre-training methods. Only SC \textit{medbertde-base-comb} outperforms all PET models with shot size 100.  
        
        \textbf{Few-shot size.} Both PET and SC models benefit from an increase in few-shot size. We observed statistical significance at shot sizes $\le$200. The smaller the shot size, the greater the relative performance gain of PET over SC models. 
        
        \textbf{Further-pretraining.} We observe notably different results for further-pretrained \textit{gbert} and \textit{medbertde} PLMs.
        
        \textit{Gbert.} PET models benefit significantly from further-pretraining with $\le$100 shots. Accuracy gradually increases with task-specific, domain-specific and combined pretraining, in that order. \textit{Gbert} SC models also benefit significantly from domain-adapted models over all shot sizes (except 10 and 400 shots), but not from task adaptation or their combination. Overall, we observe a more consistent effect of further-pretraining for PET models compared to SC models.
        
        \textit{Medbertde.} Further-pretraining shows no consistent performance improvement for  \textit{medbertde} model variants. In particular, with 20 shots, the \textit{medbertde-base} PET model outperforms the further-pretrained models, achieving a statistically significant $79.1\%$ accuracy. For few-shot sizes $10$ and $50$-$400$, the best performing model alternates between the \textit{medbertde-base-domain} and \textit{medbertde-base-comb} PET models. Similar to \textit{gbert} models, the relative gain of pretraining decreases with increasing shot sizes. It appears that our pretraining method using cardiovascular doctor's letters has no impact or may even impair the  \textit{medbertde} model. A possible reason could be that the public \textit{medbertde} model was only pretrained on $10$G of clinical and medical texts, primarily from the oncology domain. However, future research is needed for further investigation (pretraining data information cf. Fig.\ \ref{fig:methods:llm_pretraining}).
        
        \textbf{Best-performing model variant.} According to our core experiments, the overall best-performing model is the \textit{gbert} domain- and task-adapted model (\textit{gbert-base-comb}). This model achieved best accuracy scores with shot sizes $\leq$100 compared to other pretraining methods and to fine-tuned SC models with shot sizes $\le$400. When using only 20 shots, this model outperforms the SC model by $30.5$ percentage points (pp.) and the public \textit{gbert-base} PET model by $11.5$ pp. Hence, we select this model for all \textit{additional experiments}. If not further pretrained \textit{medbertde-base} outperforms public \textit{gbert-base}, this is similar to our baseline experiments. However, further-pretraining does not improve the performance of \textit{medbertde-base}, possibly due to the relatively small pretraining data size of \textit{medbertde-base} ($10$G).

        \textbf{Robustness.} Experiments were performed using three training sets and two initial random seeds. For smaller shot sizes ($\le50$ shots) standard deviation was low ($\sim2.5\%$) decreasing to less than $1\%$ for larger sizes. We observed this for \textit{gbert} and \textit{medbertde} with no impact of different pre-training methods.
     
        \subsubsection{Inspecting primary classes}
            We investigate the impact of shot size on the accuracy of predicting the selected primary section classes (Fig.\ \ref{fig:results:nocontext:primary_classes_subfigure_a}). Across shot sizes 20-50, the $F1$-scores of both classes increase in average by $9.2\%$ pp. \textit{Anamnese}, with a lower $F1$-score, benefits more from larger few-shot sizes. However, the SC model trained on the full training set significantly outperforms the 50-shot models. This is especially true for the \textit{Anamnese} class.
            Even if shot size is increased to our maximum of 400 shots, the results still differ significantly: (\textit{Anamnese}: $82.4\%$, \textit{Medikation}: 97.5\%). Results for more semi-structured classes like \textit{Medikation} are closest to the performance of the full model. For results of all shot sizes cf. Suppl. Fig. \ref{fig:suppl:results:nocontext:primary_classes_allshots}.

            \begin{figure}
                \centering
                \begin{subfigure}{0.9\textwidth}
                    \centering
                    \begin{tikzpicture}
                        \begin{axis}[
                            height=7cm,
                            ybar,
                            enlarge x limits=0.5,
                            legend style={
                            at={(0.5,-0.15)},
                            anchor=north,
                            legend columns=-1,
                            column sep=0.5em,
                            },
                            ylabel={F1-score},
                            symbolic x coords={Anamnese,Medikation},
                            xtick=data,
                            nodes near coords,
                            nodes near coords align={vertical},
                            ymax=100, % Add this line to set the maximum value of the y-axis to 100
                            ymin=0
                            ]
                            \addplot[fill=mydarkblue,nodes near coords style={rotate=90,left,color=white}] coordinates {(Anamnese,51.5) (Medikation,89.9)};
                            \addplot[fill=myroyalblue,nodes near coords style={rotate=90,left,color=white}] coordinates {(Anamnese,67.3) (Medikation,92.4)};
                            \addplot[fill=mydodgerblue,nodes near coords style={rotate=90,left,color=white}] coordinates {(Anamnese,88.1) (Medikation,98.7)};
                            \legend{20, 50, full}
                        \end{axis}
                    \end{tikzpicture}
                    \caption{}
                    \label{fig:results:nocontext:primary_classes_subfigure_a}
                \end{subfigure}
                \begin{subfigure}{0.9\textwidth}
                    \centering
                    \begin{tabular}{llp{5.2cm}}
                        \toprule
                        True label & Prediction (probability) & Shapley values \\
                        \midrule
                        Zusammenfassung & Anamnese (0.77) & {\scriptsize\colorbox{blue!87}{\color{white}Die}
                        \colorbox{blue!48}{Aufnahme}
                        \colorbox{blue!1}{der}\colorbox{blue!41}{Patientin}
                        \colorbox{blue!27}{erfolgte}\colorbox{blue!12}{bei}\colorbox{blue!33}{akutem}\colorbox{blue!92}{\color{white}Myokardinfarkt}
                        \colorbox{blue!100}{\color{white}-LRB-}\colorbox{blue!25}{STEMI}\colorbox{blue!100}{\color{white}-RRB-}} \\
                        Zusammenfassung & Zusammenfassung (0.18) & {\scriptsize\colorbox{blue!27}{Die}
                        \colorbox{red!32}{Aufnahme}
                        \colorbox{blue!32}{der}\colorbox{red!2}{Patientin}
                        \colorbox{blue!42}{erfolgte}\colorbox{blue!15}{bei}\colorbox{red!12}{akutem}\colorbox{blue!16}{Myokardinfarkt}
                        \colorbox{blue!1}{-LRB-}\colorbox{red!1}{STEMI}\colorbox{blue!47}{-RRB-}} \\
                        \bottomrule
                    \end{tabular}
                    \caption{}
                    \label{tab:results:nocontext:shap-gbert-base-comb-20-nocontext_subfigure_b}
                \end{subfigure}
                \caption{\textbf{Core experiments: primary class $F1$-score and selected Shapley values}: (a) $F1$-score scores per few-shot sizes for primary classes with using \textit{gbert-base-comb nocontext}. (b) Shapley value analysis for \textit{gbert-base-comb nocontext} with respect to \textit{Anamnese} and \textit{Zusammenfassung} prediction. First column: true label of the sample, second column: predicted label including label probability, third column: selected Shapley values. We used 20 training shots. For readability reasons, we grouped some token sequences. Further details, see Suppl. Fig. \ref{fig:suppl:results::shap-gbert-base-comb-20-nocontext}. \\Legend: \textbf{\textcolor{blue}{Blue: positive contribution}, \textcolor{red}{Red: negative contribution}}.}
                \label{fig:results:nocontext:primary_and_shap}
            \end{figure}

            While our primary classes benefit from further-pretraining, $F1$-score of \textit{Anrede} slightly decreased. A possible explanation could be that \textit{Anrede} often contains non-clinical terminology that describes a patient's place of residence, date of birth and name (Suppl.\ Fig.\ \ref{fig:suppl:discussion:f1-scores-per-label-per-pretraining}).
            
        \subsubsection{Inspecting Shapley values}
            To better understand model predictions in a few-shot setting, we further analyzed Shalpey values of the 20-shot model for the lower-performing class \textit{Anamnese}. 
            We chose a false positive sample as running example for the remainder of this study, because \textit{Anamnese} belongs to our primary classes and often suffers from a low precision rate (for 20-shots, \textit{gbert-base-comb} achieves $44.6\%$ precision and $62.2\%$  recall, cf. Suppl. Fig. \ref{fig:supp:results:confusion_matrix_nocontext}). Tab.\ \ref{tab:results:nocontext:shap-gbert-base-comb-20-nocontext_subfigure_b} illustrates selected Shapley values per token for the sample: '\textit{Die Aufnahme der Patientin erfolgte bei akutem Myokardinfarkt -LRB- STEMI -RRB- .}' towards the classes \textit{Anamnese} and \textit{Zusammenfassung},  respectively.
            
            The model incorrectly classified this sample as \textit{Anamnese},             
            with $76.8\%$ probability, while the correct class is predicted with $18.2\%$ probability score. Tokens such as \textit{Die, Aufnahme, Patient, erfolgte} positively contributed to the \textit{Anamnese} class, while the tokens \textit{Aufnahme} and \textit{Patient} negatively contributed to the correct \textit{Zusammenfassung} class. Analyzing the 20-shot training dataset, we observe that these keywords occur more frequently  in samples for \textit{Anamnese}  (\textit{Aufnahme (6x), Patient (7x), erfolgte (8x)}) than in samples from \textit{Zusammenfassung} (\textit{Aufnahme (2x), Patient (5x), erfolgte (6x)}). The token  \textit{Myokardinfarkt} positively contributes to both section classes, and to a higher extent to \textit{Anamnese}, even though we only observe this token in instances from \textit{Zusammenfassung}.
            The token sequences representing brackets \textit{-LRB-} and \textit{-RRB-} contribute strongly positively to \textit{Anamnese}. Analyzing the training data showed a higher frequency of these token in \textit{Anamnese} samples (11x) compared to \textit{Zusammenfassung} (5x).
            
            \textbf{Note on interpreting Shapley values.} Shapley values are additive: they sum up all token contributions along with the base value to yield the prediction probability. Shapley values towards different classes and of different models can not be compared by absolute value, but only relative to  other tokens for the same prediction and the same model.
            
    \subsection{Additional experiments}
        \subsubsection{Model size}
            Given the limited computational resources in clinical infrastructures, we investigated how model size affects performance and investigate its impact with finer-grained analyses. Since there is no \textit{medbertde}-large model available, we compared \textit{gbert-large} and \textit{gbert-base} models.
            
\begin{figure}
                \centering
                \begin{subfigure}{0.9\textwidth}
                    \centering
                    \begin{tabular}{lllll}
                        \toprule
                        Shot size &  SC (base) & SC (large) & PET (base) & PET (large)\\
                        \midrule
                        20 & 48.6 & 61.7 & \textbf{79.1} & 78.2 \\
                        50 & 74.6 & 81.2 & 85.6 & \textbf{86.7} \\
                        100 & 85.5 & 87.4 & 88.3 & \textbf{88.6} \\
                        400 & 90.5 & 90.7 & 89.7 & \textbf{90.4} \\
                        full & 96.7 & 96.6 & - & - \\
                        \bottomrule
                    \end{tabular}
                    \caption{}
                    \label{tab:results:additional_experiments:model_size_subfigure_a}
                \end{subfigure}
                \begin{subfigure}{0.9\textwidth}
                    \centering
                    \begin{tikzpicture}
                        \begin{axis}[
                            ybar,
                            enlarge x limits=0.5,
                            legend style={
                            at={(0.5,-0.15)},
                            anchor=north,
                            legend columns=-1,
                            column sep=0.5em,
                            },
                            ylabel={F1-score},
                            symbolic x coords={Anamnese,Medikation},
                            xtick=data,
                            nodes near coords,
                            nodes near coords align={vertical},
                            ymax=100, % Add this line to set the maximum value of the y-axis to 100
                            ymin=0
                            ]
                            \addplot[fill=mydarkblue,nodes near coords style={rotate=90,left,color=white}] coordinates {(Anamnese,51.5) (Medikation,89.8)};
                            \addplot[fill=myroyalblue,nodes near coords style={rotate=90,left,color=white}] coordinates {(Anamnese,57.4) (Medikation,90.2)};
                            \addplot[fill=mydodgerblue,nodes near coords style={rotate=90,left,color=white}] coordinates {(Anamnese,67.3) (Medikation,92.4)};
                            \addplot[fill=mynavyblue,nodes near coords style={rotate=90,left,color=white}] coordinates {(Anamnese,70.8) (Medikation,92.3)};
                            \addplot[fill=myskyblue,nodes near coords style={rotate=90,left,color=white}] coordinates {(Anamnese,88.1) (Medikation,98.7)};
                            \addplot[fill=mycornflowerblue,nodes near coords style={rotate=90,left,color=white}] coordinates {(Anamnese,87.4) (Medikation,98.8)};
                            \legend{20 base, 20 large, 50 base, 50 large, full base, full large}
                        \end{axis}
                    \end{tikzpicture}
                    \caption{}
                    \label{fig:results:additional_experiments:model_size:primary_classes_subfigure_b}
                \end{subfigure}
                \caption{\textbf{Model size}: (a) Accuracy scores for \textit{gbert-comb nocontext} PLMs using all templates on four few-shot sizes. (b) $F1$-scores for primary classes for \textit{gbert-comb no context} PLMs using all templates on various few-shot sizes.}
                \label{fig:results:additional_experiments:model_size:accuracy_primary_classes}
            \end{figure}
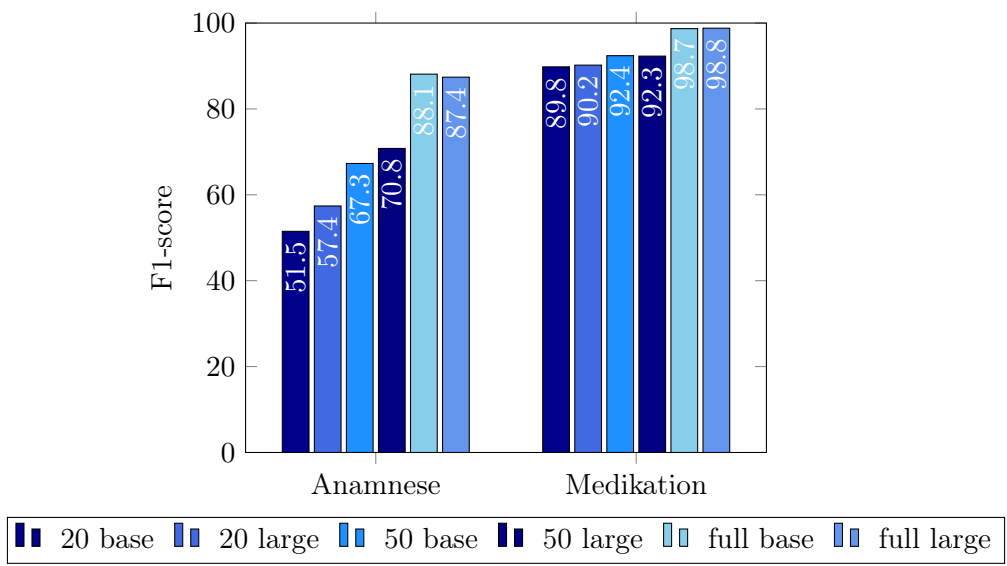

            Larger model size increases accuracy significantly, by an average of $7.2$ pp.\ for SC models $\le$ $100$. 
            PET models, by contrast, benefit less from larger model size than SC models. We even observe a slight performance decrease  for shot size 20 (Tab.\ \ref{tab:results:additional_experiments:model_size_subfigure_a}). The only significant increase, of $1.1$ points accuracy, we observed for shot size $50$.

            \textbf{Primary classes}: \textit{Gbert-large} yields an increased $F1$-score for \textit{Anamense} with both shot sizes ($20, 50$), by an average of $+4.7$ pp.\, but this is only significant for shot size $50$. By contrast, the difference in $F1$-score ($0.1\%-0.4\%$) for \textit{Medikation} is not statistically significant (Fig.\ \ref{fig:results:additional_experiments:model_size:primary_classes_subfigure_b}).
            
            \textbf{Shapley values}: Both models, \textit{gbert-base-comb} and \textit{gbert-large-comb} incorrectly classify our running example belonging to \textit{Zusammenfassung} as \textit{Anamnese}. We do not observe significant differences in the respective token contributions (Suppl.\ Fig.\ \ref{fig:suppl:results:shap-gbert-large-base-comb-20-nocontext}).
            
        \subsubsection{Null prompts}
            Inspired by insights of \cite{logan2021cutting} -- who removed all tokens from prompt templates, using null prompts instead, with comparable classification results -- we evaluated the \textit{gbert-base-comb} model using only three null prompt templates (cf. Sec.\ \ref{section:methods:templates}).
            
            Null prompts slightly decrease accuracy scores for shot sizes $\leq$50 by approximately one percentage point. For shot sizes 100 and 400 we note a slight accuracy increase. We only observed statistically significant differences in accuracy for shot-size 50 (template-based model: $85.6\%$, null-prompt model: $84.6\%$) (Suppl. Tab. \ref{tab:suppl:results:additional_experiments:null_prompts}).
            
            \textbf{Primary classes}: 
            For our primary classes we did not observe a consistent pattern. Null prompts have a slightly negative impact on $F1$-scores for \textit{Anamnese} and \textit{Medikation} with 20 shots. By contrast, with 50 shots, 
            accuracy significantly decreases for \textit{Anamnese}, but slightly increases for \textit{Medikation} ($92.4\%$ vs.\ $95.9\%$).

        \subsubsection{Adding context}
           Predicting section classes is difficult for tokens that frequently occur in different classes, as discussed for the example in Fig.\ \ref{fig:results:nocontext:primary_and_shap}.
           To reduce the degree of ambiguity of individual tokens, we experimented with two types of \textit{contextualization} of classification instances: Adding (1) the previous and subsequent paragraph (\textit{context}) and (2) only the previous paragraph (\textit{prevcontext}). Suppl. Fig.\ \ref{fig:suppl:results:additional_experiments:context} shows that across all few-shot sizes, (1) \textit{context} 
           (with mean $+2.4$ accuracy points) and (2) \textit{prevcontext} (with mean $+1.6$ accuracy points) both achieve significantly higher accuracy than \textit{nocontext} models (cf. Sec.\ \ref{section:methods:additional_experiments}).
           
            \textbf{Primary classes}: \textit{Context} models improve the $F1$-scores for both primary classes (by mean $+7.8$ points for \textit{Anamnese} and $+1.3$ for \textit{Medikation})  (Fig.\ \ref{fig:results:context:primary_classes_subfigure_a}). For \textit{Anamnese},  statistically significant improvement is only reached using 50 shots.

            \begin{figure}
                \centering
                \begin{subfigure}{0.9\textwidth}
                    \centering
                    \begin{tikzpicture}[scale=0.8]
                        \begin{axis}[
                            ybar,
                            enlarge x limits=0.5,
                            legend style={
                            at={(0.5,-0.15)},
                            anchor=north,
                            legend columns=3,
                            column sep=0.5em, 
                            },
                            ylabel={F1-score},
                            symbolic x coords={Anamnese,Medikation},
                            xtick=data,
                            nodes near coords,
                            nodes near coords align={vertical},
                            ymax=100, % Add this line to set the maximum value of the y-axis to 100
                            ymin=0
                            ]
                            \addplot[fill=mydarkblue,nodes near coords style={rotate=90,left,color=white}] coordinates {(Anamnese,51.5) (Medikation,89.8)};
                            \addplot[fill=myroyalblue,nodes near coords style={rotate=90,left,color=white}] coordinates {(Anamnese,56.6) (Medikation,90.8)};
                            \addplot[fill=mydodgerblue,nodes near coords style={rotate=90,left,color=white}] coordinates {(Anamnese,67.3) (Medikation,92.4)};
                            \addplot[fill=mynavyblue,nodes near coords style={rotate=90,left,color=white}] coordinates {(Anamnese,77.8) (Medikation,94)};
                            \addplot[fill=myskyblue,nodes near coords style={rotate=90,left,color=white}] coordinates {(Anamnese,88.1) (Medikation,98.7)};
                            \addplot[fill=mycornflowerblue,nodes near coords style={rotate=90,left,color=white}] coordinates {(Anamnese,96.8) (Medikation,99.3)};
                            \legend{20 nocontext, 20 context, 50 nocontext, 50 context, full nocontext, full context}
                        \end{axis}
                    \end{tikzpicture}
                    \caption{}
                    \label{fig:results:context:primary_classes_subfigure_a}
                \end{subfigure}
                \begin{subfigure}{0.9\textwidth}
                    \centering
                    \begin{tabular}{llp{5.2cm}}
                        \toprule
                        True label & Prediction (probability) & Shapley values \\
                        \midrule
                        Zusammenfassung & Zusammenfassung (0.18) & {\scriptsize\colorbox{blue!27}{Die}
                        \colorbox{red!32}{Aufnahme}
                        \colorbox{blue!32}{der}\colorbox{red!2}{Patientin}
                        \colorbox{blue!42}{erfolgte}\colorbox{blue!15}{bei}\colorbox{red!12}{akutem}\colorbox{blue!16}{Myokardinfarkt}
                        \colorbox{blue!1}{-LRB-}\colorbox{red!1}{STEMI}\colorbox{blue!47}{-RRB-}} \\
                        Zusammenfassung & Zusammenfassung (0.87) & {\scriptsize\colorbox{blue!57}{\color{white}PREVIOUS CONTEXT TOKENS}
                        \colorbox{white}{[SEP]}\colorbox{blue!1}{Die Aufnahme der}
                        \colorbox{blue!0.7}{Patientin erfolgte bei}
                        \colorbox{blue!17}{akutem}\colorbox{blue!24}{Myokardinfarkt}
                        \colorbox{blue!12}{-LRB-}\colorbox{blue!13}{STEMI}\colorbox{blue!32}{-RRB-}\colorbox{white}{[SEP]}
                        \colorbox{blue!100}{\color{white}SUBSEQUENT CONTEXT TOKENS}} \\
                        \bottomrule
                    \end{tabular}
                    \caption{}
                    \label{tab:results:context:shap-gbert-base-comb-20-nocontext_context_subfigure_b}
                \end{subfigure}
                \caption{\textbf{Additional experiments (context) - primary classes $F1$-scores and selected Shapley values}: (a) $F1$-scores per few-shot sizes for primary classes with \textit{nocontext} and \textit{context} using \textit{gbert-base-comb}. Comparing to \textit{gbert-base-comb} trained on full training data with \textit{nocontext} and \textit{context}. (b) Shapley value analysis for \textit{gbert-base-comb nocontext} and \textit{gbert-base-comb context}. First column: true label of the sample, second column: predicted label including label probability, third column: selected Shapley values. We used 20 training shots. For readability reasons, we grouped some token sequences. Further details, see Suppl. Fig. \ref{fig:suppl:results:shap-gbert-base-comb-20-context_nocontext}.\\Legend: \textbf{\textcolor{blue}{Blue: positive contribution}, \textcolor{red}{Red: negative contribution}}. }
                \label{fig:results:context:primary_and_shap}
            \end{figure}
            
            \textbf{Shapley values}: \textit{gbert-base-comb context} correctly classifies our running example with $86.6\%$ probability (Tab.\ \ref{tab:results:context:shap-gbert-base-comb-20-nocontext_context_subfigure_b}). Most highly contributing tokens belong to the context (previous or following, with Shapley values: $0.057+0.596$), while the main paragraph has an accumulated Shapley value of $0.106$. 
            The previous context contains the sequence: \textit{Zusammenfassende Beurteilung}, a frequent section-specific title. The subsequent paragraph is the longest paragraph ($37$ tokens). Previously negatively contributing tokens (\textit{Aufnahme} and \textit{Patient}) are now positively contributing to the correct class: \textit{Zusammenfassung}.

        \subsubsection{Combining best-performing methods}
            Our \textit{core experiments} indicated that the \textit{gbert-base-comb} model performed best of all tested models. The \textit{additional experiments} showed that models using all five templates (cf.\ Section \ref{section:methods:templates}), a BERT-large architecture and contextualization often achieved the best performance. Hence, we investigated whether this combination (\textit{gbert-large-comb context} trained with all templates) could further close the performance gap to a model trained on full training set.
            
            Tab.\ \ref{tab:results:additional_experiments:combined_methods} shows that \textit{gbert-large-comb context} significantly outperforms
            both \textit{gbert-base-comb} and \textit{gbert-large-comb} without context. Moreover, \textit{gbert-large-comb context} statistically significantly outperforms \textit{gbert-base-comb context} for $20, 100$ and $400$ shots. Overall, the \textit{gbert-large-comb context} outperforms \textit{nocontext} and \textit{base} models over all shot-sizes, yielding best results with \textit{400 shots}. Yet, PET still lags behind the \textit{full} SC setting, with a minimal gap of $-5.2$ points accuracy.
            
            \textbf{Primary classes}: For our primary classes, \textit{gbert-large-comb context} now outperforms \textit{gbert-large-comb nocontext} by large margin (Fig.\ \ref{fig:additional_experiments:combined_methods:primary_classes_subfigure_a}). Only the 50-shot results for \textit{Anamnese} are not statistically significant ($F1$-score of all shot-sizes cf. Suppl. Fig. \ref{fig:suppl:results:context:primary_classes_allshots}).

            \begin{figure}
                \centering
                \begin{subfigure}{0.9\textwidth}
                    \centering
                   \begin{tikzpicture}[scale=0.7]
                        \begin{axis}[
                            ybar,
                            enlarge x limits=0.5,
                            legend style={
                            at={(0.5,-0.15)},
                            anchor=north,
                            legend columns=-1,
                            column sep=0.5em,
                            },
                            ylabel={F1-score},
                            symbolic x coords={Anamnese,Medikation},
                            xtick=data,
                            nodes near coords,
                            nodes near coords align={vertical},
                            ymax=100, % Add this line to set the maximum value of the y-axis to 100
                            ymin=0
                            ]
                            \addplot[fill=mydarkblue,nodes near coords style={rotate=90,left,color=white}] coordinates {(Anamnese,57.4) (Medikation,90.2)};
                            \addplot[fill=myroyalblue,nodes near coords style={rotate=90,left,color=white}] coordinates {(Anamnese,71.1) (Medikation,93.0)};
                            \addplot[fill=mydodgerblue,nodes near coords style={rotate=90,left,color=white}] coordinates {(Anamnese,70.8) (Medikation,92.3)};
                            \addplot[fill=mynavyblue,nodes near coords style={rotate=90,left,color=white}] coordinates {(Anamnese,80.4) (Medikation,93.8)};
                            \addplot[fill=myskyblue,nodes near coords style={rotate=90,left,color=white}] coordinates {(Anamnese,96.6) (Medikation,99.5)};
                            \legend{20 nocontext, 20 context, 50 nocontext, 50 context, full context}
                        \end{axis}
                    \end{tikzpicture}
                    \caption{}        \label{fig:additional_experiments:combined_methods:primary_classes_subfigure_a}
                \end{subfigure}
                \begin{subfigure}{0.9\textwidth}
                        \centering
                        \begin{tabular}{llp{5.2cm}}
                            \toprule
                            True label & Prediction (probability) & Shapley values \\
                            \midrule
                            Zusammenfassung & Zusammenfassung (0.87) & {\scriptsize\colorbox{blue!57}{\color{white}PREVIOUS CONTEXT TOKENS}
                            \colorbox{white}{[SEP]}\colorbox{blue!1}{Die Aufnahme der}
                            \colorbox{blue!0.7}{Patientin erfolgte bei}
                            \colorbox{blue!17}{akutem}\colorbox{blue!24}{Myokardinfarkt}
                            \colorbox{blue!12}{-LRB-}\colorbox{blue!13}{STEMI}\colorbox{blue!32}{-RRB-}\colorbox{white}{[SEP]}
                            \colorbox{blue!100}{\color{white}SUBSEQUENT CONTEXT TOKENS}} \\
                            Zusammenfassung & Zusammenfassung (0.99) & {\scriptsize\colorbox{red!6}{PREVIOUS CONTEXT TOKENS}
                            \colorbox{white}{[SEP]}\colorbox{blue!100}{\color{white}Die Aufnahme der}
                            \colorbox{blue!70}{\color{white}Patientin erfolgte bei}
                            \colorbox{blue!25}{akutem}\colorbox{blue!33}{Myokardinfarkt}
                            \colorbox{blue!11}{-LRB-}\colorbox{blue!4}{STEMI}\colorbox{red!8}{-RRB-}\colorbox{white}{[SEP]}
                            \colorbox{blue!100}{\color{white}SUBSEQUENT CONTEXT TOKENS}} \\
                            \bottomrule
                        \end{tabular}
                        \caption{}\label{tab:results:combined_methods:primary_and_shap_subfigure_b}
                \end{subfigure}
                \caption{\textbf{Additional experiments (combined methods) - primary classes $F1$-scores and selected Shapley values}: (a) $F1$-scores per few-shot sizes for primary classes with \textit{nocontext} and \textit{context} using \textit{gbert-large-comb}. Comparing to \textit{gbert-large-comb} trained on full training data with \textit{context}. (b) Shapley value analysis for \textit{gbert-base-comb context} and \textit{gbert-large-comb context}. First column: true label of the sample, second column: predicted label including label probability, third column: selected Shapley values. We used 20 training shots. For readability reasons, we grouped some token sequences. More detailed results, see Suppl. Fig. \ref{fig:suppl:results:additional_experiments:combined_methods:shap-gbert-base-large-comb-20-context}.\\Legend: \textbf{\textcolor{blue}{Blue: positive contribution}, \textcolor{red}{Red: negative contribution}}. }
                \label{fig:results:combined_methods:primary_and_shap}
            \end{figure}
            
            We also compared the \textit{large} and \textit{base} versions of \textit{gbert-*-comb context}. The $F1$-score for \textit{Anamnese} is significantly increased by $+14.6$ points with 20 shots, and by $+2.6$ points with 50 shots. Performance for \textit{Medikation} is significantly increased by $+2.2$ points with 20 shots, but insignificantly decreased with 50 shots. (Suppl.\ Fig.\ \ref{fig:suppl:results:additional_experiments:combined_methods:primary_classes:nocontext_vs_context})
            
            \textbf{Shapley values}: We tested whether the token contributions differ between the \textit{large} and \textit{base} \textit{gbert-*-comb context} models (Tab.\ \ref{tab:results:combined_methods:primary_and_shap_subfigure_b}). The large model predicts the true class \textit{Zusammenfassung} with $99.2\%$ probability, $+12.7$ points above the \textit{base} model.
            The \textit{large} context model now also places greater emphasis on the main paragraph, as opposed to the context. The ratio of the accumulated Shapley values ($\frac{{classified\ instance}}{{context\ paragraphs}}$, higher is better) is $0.36$ for \textit{gbert-large-comb context}
            and $0.16$ for \textit{gbert-base-comb context}.
            
\begin{table}
    \centering
    \caption{\textbf{Combining and evaluating best performing methods}: Accuracy scores for \textit{gbert-large-comb context} evaluated on few-shot sizes [$20,50,100,400$] with \textit{base vs.\ large} model sizes in \textit{context vs.\ nocontext} settings using PET. Comparison to corresponding SC model fine-tuned on full training set.}
    \begin{tabular}{lllll}
        \toprule
        Shot size & Base \textit{nocontext} & Large \textit{nocontext} & Base \textit{context} & \textbf{Large  \textit{context}}\\
        \midrule
        20 & 79.1 & 78.2 & 80.5 & \textbf{84.3} \\
        50 & 85.6 & 86.7 & 89.2 & \textbf{89.4} \\
        100 & 88.3 & 88.6 & 90.9 & \textbf{91.3} \\
        400 & 90 & 90.4 & 92.8 & \textbf{93.4} \\
        \midrule
        full (SC) & 96.7 & 96.6 & 98.6 & 98.6  \\
        \bottomrule
    \end{tabular}    \label{tab:results:additional_experiments:combined_methods}
\end{table}

 \section{Discussion}
    \label{section:discussion}
    In this section, we discuss our empirical findings in light of the challenges and proposed solutions outlined in Section \ref{intro}.
    \begin{enumerate}[label=S$_\arabic*$]
        \item \textbf{Domain- and Expert-dependent.}
        In in-depth evaluations we compared four pretraining approaches using PET and SC for two public German-language models \citep{gururangan2020don}: 
        (1) \textit{initial pre-training} using general German texts with \textit{gbert} vs.\ exclusively medical and clinical data with \textit{medbertde} (Suppl. Fig. \ref{fig:methods:llm_pretraining}); and \textit{further-pretraining} of these PLMs for (2) \textit{task-adaption}, (3) \textit{domain-adaptation} and (4) combined \textit{task and domain-adaptation}. 
        
        \textbf{Finding.}  \textit{Gbert} overall accuracy gradually improved with further-pretraining. The task- and domain-adapted \textit{gbert-base-comb} performs best compared to all models, and with only \textit{20 shots} outperforms \textit{gbert-base} by $+11.5$ accuracy points. Also, the positive effect of further-pretraining was more consistent for PET compared to SC models. 
        By contrast, further-pretrained \textit{medbertde}-based SC and PET models did not achieve consistent performance improvements.
        
        \textbf{Finding.} 
        Pre-training from scratch with sufficient clinical and medical data can benefit various MIE tasks. However, when pretraining data is limited and/or concentrated on a narrow domain, e.g., oncology, as in the case of \textit{medbertde}, further-pretraining was found not to enhance performance.
        
        \textbf{Finding.}  While \textit{medbertde-base} without further-pretraining outperformed \textit{gbert-base} in all shot sizes, and similarly when trained on the \textit{full} dataset (Fig.\ \ref{fig:results:core_experiments}), it did not improve performance if further pre-trained and was outperformed by further-pretrained \textit{gbert-base}.

        \item \textbf{Resource-constraints.}
        Prompt-based fine-tuning with PET produces superior classification results in few-shot learning scenarios. 
        
        \textbf{Finding.} We observed a steady increase in the performance of PET compared to SC models With decreasing few-shot training set sizes (400-10 shots). Using 20 shots, the PET \textit{gbert-base-comb nocontext} model outperforms the corresponding SC model by $+30.5$ pp. The same \textit{gbert-base-comb nocontext} PET model with 50 shots even rivals the SC model trained on \textit{full} data, leaving a gap of $-11.1$ pp. 
        Especially semi-structured section classes, such as \textit{Medikation}, perform close to the full model by $-6.3$ pp.\ (Fig.\ \ref{fig:results:nocontext:primary_classes_subfigure_a}). Our few-shot models are also \textit{robust} as measured by standard deviation.
        
        \textbf{Finding.} \textit{Null prompts} exhibit comparable results with no significant difference in performance, especially with few-shot sizes exceeding $100$.
        
        \textbf{Finding.} Contextualize data with surrounding \textit{context} paragraphs improved classification results for most section classes, especially primary classes. It allowed our base models to correctly predict our running false-positive sample as \textit{Zusammenfassung}. However, compared to the base models interpretability analysis using SHAP revealed that the large model places greater emphasis on main paragraph tokens rather than on context paragraphs. Contextualization further reduced the accuracy gap between \textit{gbert-*-comb context}-based PET models trained on 50 shots to the \textit{full} SC model to $-9$ to $-9.5$pp; for classes such as \textit{Medikation} even to $-5$ to $-6$pp. Contextualization does not require complex pre-processing or manual annotation.

        \item \textbf{On-premise}:
        Using smaller models saves computational resources. We therefore compared classification performances of \textit{base} and \textit{large} BERT PLMs.
        
        \textbf{Finding.} Large PLMs  achieve better classification results. However, model size has a lower impact on the performance of PET compared to SC models (Fig. \ref{tab:results:additional_experiments:model_size_subfigure_a}). For classes such as \textit{Medikation} the further-pretrained \textit{gbert-base-comb} PLM performs almost on par with \textit{gbert-large-comb}  (Fig.\ \ref{fig:results:additional_experiments:model_size:primary_classes_subfigure_b}).

        \textbf{Finding.} For complex sections with free text such as \textit{Anamnese}, \textit{gbert-large} PLMs achieved better performance. They also better recognize contextualized instances (Tab.\ \ref{tab:results:combined_methods:primary_and_shap_subfigure_b} and Suppl. Sect. \ref{sect:suppl:ablation_tests:SEP_recognition}).
        
        \item \textbf{Transparency.}
        Shapley values \citep{Lundberg_NIPS2017_7062}, an interpretability method based on  saliency features, helped identify problems in \textit{training data quality} and \textit{model decisions}. We identified tokens that frequently occur in false-positive classes by analyzing model predictions (Fig. \ref{fig:results:context:primary_and_shap}). 
        
        \textbf{Finding.} The use of Shapley features is especially beneficial in few-shot scenarios, as it enables data engineers to select few-shot samples with high precision.
        
        Shapley values also proved instrumental for identifying problems with contextualization: It became clear that with very small shot sizes, and for section classes with short spans, the model prioritized the context over the instance to be classified. They also provided evidence that our \textit{gbert-large-comb} model outperforms its base counterpart by focusing on key parts of contextualized samples.
        
        \textbf{Finding.} Our analysis of Shapley values showed that \textit{gbert-large-comb} makes more reliable predictions than \textit{gbert-base-comb}, by prioritizing features of instances to be classified over context (Tab.\ \ref{tab:results:combined_methods:primary_and_shap_subfigure_b}).
        
    \end{enumerate}

\section{Conclusions \& Recommendations}

    In this work we have presented best-practice strategies to identify an ideal setup to address the multi-faceted challenges of conducting a MIE task, such as clinical section classification, in a clinical setting for low-resource languages.
    
    To reduce the demand for clinical knowledge in MIE we showed in S$_1$ that few-shot prompting performed particularly well with further-pretrained general-domain PLMs, and helped to reduce the demand of clinical expert knowledge for manual data annotation. Our experiments  revealed that pre-training data has a strong impact on few-shot learning results (see S$_2$), especially if training data is limited. Specifically, \textit{general domain PLMs} such as \textit{gbert}, pre-trained on massive amounts of general language, can be effectively domain- and task-adapted by \textit{further-pretraining} on clinical routine data. To the contrary, PLMs pre-trained on domain-specific data from scratch, such as \textit{medbertde} may outperform \textit{gbert} if not further-pretrained, but may not benefit from further-pretraining. Therefore, if further-pretraining for domain adaptation is not feasible due to IT constraints, we recommend choosing clinical PLMs like \textit{medbertde} over non-adapted general PLMs.

     Our study indicated that prompt-based learning methods improve classification results if annotated data is rare, and effectively reduces time investment and costs of manual data annotation. The larger the amount of annotated data, the higher the efficiency of \textit{null prompts},  which further save 
     engineering time (see S$_2$). Moreover, contextualizing classification instances improves performance, especially for the primary classes, and further closes the gap to \textit{full} models.
    
    We found in S$_3$ that in case of limited computing resources,   prompting methods allow practitioners to employ smaller PLMs in a few-shot scenario, while  achieving classification results comparable to larger models. However, free-text sections, such as \textit{Anamnese} may still benefit from larger model architectures (Fig.\ \ref{fig:results:additional_experiments:model_size:accuracy_primary_classes}).
    
    Finally, in S$_4$, we addressed the need for transparent and trustworthy model predictions in low-resource clinical NLP, and possible use cases for \textit{interpretability} methods. Our study demonstrates that the analysis of Shapley values can help improve training data quality, which is especially important with small shot sizes. Examining Shapley values, or similar interpretability methods, can also inform model selection, by revealing tokens that contribute to classification errors in specific model types. Finally, model interpretability is crucial in safety-critical domains such as clinical routine, to enhance the trustworthiness of model predictions.

    Our study presents strategies for optimising MIE in low-resource clinical language settings. 
    It highlights the benefits of few-shot prompting with further-pretrained PLMs as a measure to reduce the demand for manual annotation by clinicians. We further demonstrate that prompt-based learning and contextualisation significantly enhance classification accuracy, especially in low-resource scenarios, while keeping demands on computing resources low.
    We are certain that these insights help to advance MIE tasks in clinical settings in the context of low-resource languages such as German.

\printbibliography

\newpage

\appendix
\section*{Supplementary information}
    \renewcommand\thefigure{S\arabic{figure}}
    \renewcommand\thetable{S\arabic{table}}
    \renewcommand{\thesubsection}{S\arabic{subsection}}
    \setcounter{figure}{0}
    \setcounter{table}{0}
    \setcounter{subsection}{0}
    \FloatBarrier

    \subsection{Ablation test}
        \subsubsection{Comparing to \textit{medbertde}}
            \label{sect:suppl:ablation_tests:medbert}
            While \textit{gbert-base-comb nocontext} was the overall best-performing model in our core experiments, the publicly available \textit{medbertde-base nocontext} pretrained on medical data from scratch achieved superior results in frequent scenarios. Hence, we assessed, how \textit{medbertde-base context} performs in comparison to our final model \textit{gbert-large-comb context}. We trained a \textit{medbertde-base context} model without further pretraining, as further pretraining did not show a consistent performance improvement in the core experiments. For 20 shots \textit{medbertde-base context} achieved statistically significant better accuracy results than \textit{gbert-large-comb context} ($86.2\%$ vs. $84.3\%$). Performance differences for 50 and 100 shots are not significant, while using 400 shots, \textit{gbert-large-comb context} achieves better results ($93.4\%$ vs. $92.4\%$) (Suppl. Tab.\ \ref{tab:suppl:results:additional_experiments:medbertde}).\\
            \textbf{Primary classes}: With regard to the primary classes, the $F1$-score of \textit{gbert-large-comb context} is significantly better than \textit{medbertde-base context} for the \textit{Anamnese} class with mean $+4.2$ percentage points. This support our hypothesis, that larger PLMs are superior on complex free text section classes (cf. Section \ref{section:discussion}). To a lesser extent, but significantly, \textit{medbertde-base context} achieves better $F1$-scores for the \textit{Medikation} class. (Suppl. Fig. \ref{fig:suppl:results:additional_experiments:combined_methods:primary_classes:medbertde})
        \subsubsection{Inspecting \texttt{[SEP]} recognition}
        \label{sect:suppl:ablation_tests:SEP_recognition}
            We observed significant performance drops for classes such as: \textit{AllergienUnverträglichkeitenRisiken, Anrede} and \textit{Mix}. To gain better understanding of this decline, (1) we performed \textit{fine-grained class analysis} for samples from \textit{Anrede} and (2) \textit{analyzed Shapley values}. (1) We found that precision dropped from $98.2\%$ to $48.1\%$. $99$ out of $131$ instances were misclassified as \textit{Diagnosen}. Even if we use 400 training shots, the \textit{gbert-base-comb context} model still achieves a low precision rate ($56.8\%$). This can only be improved to $68.3\%$ using a \textit{gbert-large-comb context} model. Both precision scores are significantly below \textit{nocontext} models with a precision of $98.2\%$.(2) Shapley values shed further light on typical patterns of this section samples. The three classes \textit{AllergienUnverträglichkeitenRisiken, Anrede, Mix} typically contain only a single paragraph or sentence. \textit{Anrede} paragraph are typically followed by a \textit{Diagnosen} paragraph, containing section headers such as \textit{Aktuelle Diagnosen:}.
            
\begin{figure}
        \centering
        % This file was created with tikzplotlib v0.10.1.
\begin{tikzpicture}

\definecolor{darkslategray38}{RGB}{38,38,38}
\definecolor{lavender234234242}{RGB}{234,234,242}

\begin{axis}[
axis background/.style={fill=lavender234234242},
axis line style={white},
colorbar,
colorbar style={ylabel={}},
colormap={mymap}{[1pt]
  rgb(0pt)=(0.968627450980392,0.984313725490196,1);
  rgb(1pt)=(0.870588235294118,0.92156862745098,0.968627450980392);
  rgb(2pt)=(0.776470588235294,0.858823529411765,0.937254901960784);
  rgb(3pt)=(0.619607843137255,0.792156862745098,0.882352941176471);
  rgb(4pt)=(0.419607843137255,0.682352941176471,0.83921568627451);
  rgb(5pt)=(0.258823529411765,0.572549019607843,0.776470588235294);
  rgb(6pt)=(0.129411764705882,0.443137254901961,0.709803921568627);
  rgb(7pt)=(0.0313725490196078,0.317647058823529,0.611764705882353);
  rgb(8pt)=(0.0313725490196078,0.188235294117647,0.419607843137255)
},
point meta max=2988,
point meta min=0,
tick align=outside,
title={Confusion Matrix},
x grid style={white},
xlabel=\textcolor{darkslategray38}{Predicted labels},
xmajorticks=true,
xmin=0, xmax=9,
xtick style={color=darkslategray38},
xtick={0.5,1.5,2.5,3.5,4.5,5.5,6.5,7.5,8.5},
xticklabel style={rotate=90.0},
xticklabels={
  Abschluss,
  AllergienUnv.,
  Anamnese,
  Anrede,
  Befunde,
  Diagnosen,
  Medikation,
  Mix,
  Zusammenfassung
},
y dir=reverse,
y grid style={white},
ylabel=\textcolor{darkslategray38}{True labels},
ymajorticks=true,
ymin=0, ymax=9,
ytick style={color=darkslategray38},
ytick={0.5,1.5,2.5,3.5,4.5,5.5,6.5,7.5,8.5},
yticklabels={
  Abschluss,
  AllergienUnv.,
  Anamnese,
  Anrede,
  Befunde,
  Diagnosen,
  Medikation,
  Mix,
  Zusammenfassung
}
]
\addplot graphics [includegraphics cmd=\pgfimage,xmin=0, xmax=9, ymin=9, ymax=0] {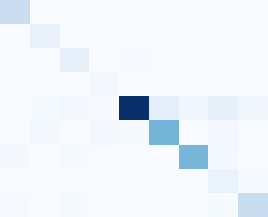};
\draw (axis cs:0.5,0.5) node[
  scale=0.6,
  text=darkslategray38,
  rotate=0.0
]{689};
\draw (axis cs:1.5,0.5) node[
  scale=0.6,
  text=darkslategray38,
  rotate=0.0
]{0};
\draw (axis cs:2.5,0.5) node[
  scale=0.6,
  text=darkslategray38,
  rotate=0.0
]{0};
\draw (axis cs:3.5,0.5) node[
  scale=0.6,
  text=darkslategray38,
  rotate=0.0
]{1};
\draw (axis cs:4.5,0.5) node[
  scale=0.6,
  text=darkslategray38,
  rotate=0.0
]{0};
\draw (axis cs:5.5,0.5) node[
  scale=0.6,
  text=darkslategray38,
  rotate=0.0
]{0};
\draw (axis cs:6.5,0.5) node[
  scale=0.6,
  text=darkslategray38,
  rotate=0.0
]{0};
\draw (axis cs:7.5,0.5) node[
  scale=0.6,
  text=darkslategray38,
  rotate=0.0
]{1};
\draw (axis cs:8.5,0.5) node[
  scale=0.6,
  text=darkslategray38,
  rotate=0.0
]{4};
\draw (axis cs:0.5,1.5) node[
  scale=0.6,
  text=darkslategray38,
  rotate=0.0
]{0};
\draw (axis cs:1.5,1.5) node[
  scale=0.6,
  text=darkslategray38,
  rotate=0.0
]{222};
\draw (axis cs:2.5,1.5) node[
  scale=0.6,
  text=darkslategray38,
  rotate=0.0
]{0};
\draw (axis cs:3.5,1.5) node[
  scale=0.6,
  text=darkslategray38,
  rotate=0.0
]{0};
\draw (axis cs:4.5,1.5) node[
  scale=0.6,
  text=darkslategray38,
  rotate=0.0
]{0};
\draw (axis cs:5.5,1.5) node[
  scale=0.6,
  text=darkslategray38,
  rotate=0.0
]{8};
\draw (axis cs:6.5,1.5) node[
  scale=0.6,
  text=darkslategray38,
  rotate=0.0
]{0};
\draw (axis cs:7.5,1.5) node[
  scale=0.6,
  text=darkslategray38,
  rotate=0.0
]{6};
\draw (axis cs:8.5,1.5) node[
  scale=0.6,
  text=darkslategray38,
  rotate=0.0
]{0};
\draw (axis cs:0.5,2.5) node[
  scale=0.6,
  text=darkslategray38,
  rotate=0.0
]{1};
\draw (axis cs:1.5,2.5) node[
  scale=0.6,
  text=darkslategray38,
  rotate=0.0
]{7};
\draw (axis cs:2.5,2.5) node[
  scale=0.6,
  text=darkslategray38,
  rotate=0.0
]{246};
\draw (axis cs:3.5,2.5) node[
  scale=0.6,
  text=darkslategray38,
  rotate=0.0
]{6};
\draw (axis cs:4.5,2.5) node[
  scale=0.6,
  text=darkslategray38,
  rotate=0.0
]{12};
\draw (axis cs:5.5,2.5) node[
  scale=0.6,
  text=darkslategray38,
  rotate=0.0
]{2};
\draw (axis cs:6.5,2.5) node[
  scale=0.6,
  text=darkslategray38,
  rotate=0.0
]{1};
\draw (axis cs:7.5,2.5) node[
  scale=0.6,
  text=darkslategray38,
  rotate=0.0
]{0};
\draw (axis cs:8.5,2.5) node[
  scale=0.6,
  text=darkslategray38,
  rotate=0.0
]{6};
\draw (axis cs:0.5,3.5) node[
  scale=0.6,
  text=darkslategray38,
  rotate=0.0
]{0};
\draw (axis cs:1.5,3.5) node[
  scale=0.6,
  text=darkslategray38,
  rotate=0.0
]{0};
\draw (axis cs:2.5,3.5) node[
  scale=0.6,
  text=darkslategray38,
  rotate=0.0
]{0};
\draw (axis cs:3.5,3.5) node[
  scale=0.6,
  text=darkslategray38,
  rotate=0.0
]{99};
\draw (axis cs:4.5,3.5) node[
  scale=0.6,
  text=darkslategray38,
  rotate=0.0
]{0};
\draw (axis cs:5.5,3.5) node[
  scale=0.6,
  text=darkslategray38,
  rotate=0.0
]{0};
\draw (axis cs:6.5,3.5) node[
  scale=0.6,
  text=darkslategray38,
  rotate=0.0
]{0};
\draw (axis cs:7.5,3.5) node[
  scale=0.6,
  text=darkslategray38,
  rotate=0.0
]{0};
\draw (axis cs:8.5,3.5) node[
  scale=0.6,
  text=darkslategray38,
  rotate=0.0
]{0};
\draw (axis cs:0.5,4.5) node[
  scale=0.6,
  text=darkslategray38,
  rotate=0.0
]{4};
\draw (axis cs:1.5,4.5) node[
  scale=0.6,
  text=darkslategray38,
  rotate=0.0
]{26};
\draw (axis cs:2.5,4.5) node[
  scale=0.6,
  text=darkslategray38,
  rotate=0.0
]{84};
\draw (axis cs:3.5,4.5) node[
  scale=0.6,
  text=darkslategray38,
  rotate=0.0
]{21};
\draw (axis cs:4.5,4.5) node[
  scale=0.6,
  text=white,
  rotate=0.0
]{2988};
\draw (axis cs:5.5,4.5) node[
  scale=0.6,
  text=darkslategray38,
  rotate=0.0
]{284};
\draw (axis cs:6.5,4.5) node[
  scale=0.6,
  text=darkslategray38,
  rotate=0.0
]{121};
\draw (axis cs:7.5,4.5) node[
  scale=0.6,
  text=darkslategray38,
  rotate=0.0
]{266};
\draw (axis cs:8.5,4.5) node[
  scale=0.6,
  text=darkslategray38,
  rotate=0.0
]{120};
\draw (axis cs:0.5,5.5) node[
  scale=0.6,
  text=darkslategray38,
  rotate=0.0
]{0};
\draw (axis cs:1.5,5.5) node[
  scale=0.6,
  text=darkslategray38,
  rotate=0.0
]{62};
\draw (axis cs:2.5,5.5) node[
  scale=0.6,
  text=darkslategray38,
  rotate=0.0
]{2};
\draw (axis cs:3.5,5.5) node[
  scale=0.6,
  text=darkslategray38,
  rotate=0.0
]{99};
\draw (axis cs:4.5,5.5) node[
  scale=0.6,
  text=darkslategray38,
  rotate=0.0
]{53};
\draw (axis cs:5.5,5.5) node[
  scale=0.6,
  text=darkslategray38,
  rotate=0.0
]{1428};
\draw (axis cs:6.5,5.5) node[
  scale=0.6,
  text=darkslategray38,
  rotate=0.0
]{7};
\draw (axis cs:7.5,5.5) node[
  scale=0.6,
  text=darkslategray38,
  rotate=0.0
]{86};
\draw (axis cs:8.5,5.5) node[
  scale=0.6,
  text=darkslategray38,
  rotate=0.0
]{1};
\draw (axis cs:0.5,6.5) node[
  scale=0.6,
  text=darkslategray38,
  rotate=0.0
]{75};
\draw (axis cs:1.5,6.5) node[
  scale=0.6,
  text=darkslategray38,
  rotate=0.0
]{0};
\draw (axis cs:2.5,6.5) node[
  scale=0.6,
  text=darkslategray38,
  rotate=0.0
]{49};
\draw (axis cs:3.5,6.5) node[
  scale=0.6,
  text=darkslategray38,
  rotate=0.0
]{0};
\draw (axis cs:4.5,6.5) node[
  scale=0.6,
  text=darkslategray38,
  rotate=0.0
]{4};
\draw (axis cs:5.5,6.5) node[
  scale=0.6,
  text=darkslategray38,
  rotate=0.0
]{8};
\draw (axis cs:6.5,6.5) node[
  scale=0.6,
  text=darkslategray38,
  rotate=0.0
]{1412};
\draw (axis cs:7.5,6.5) node[
  scale=0.6,
  text=darkslategray38,
  rotate=0.0
]{75};
\draw (axis cs:8.5,6.5) node[
  scale=0.6,
  text=darkslategray38,
  rotate=0.0
]{4};
\draw (axis cs:0.5,7.5) node[
  scale=0.6,
  text=darkslategray38,
  rotate=0.0
]{6};
\draw (axis cs:1.5,7.5) node[
  scale=0.6,
  text=darkslategray38,
  rotate=0.0
]{7};
\draw (axis cs:2.5,7.5) node[
  scale=0.6,
  text=darkslategray38,
  rotate=0.0
]{0};
\draw (axis cs:3.5,7.5) node[
  scale=0.6,
  text=darkslategray38,
  rotate=0.0
]{0};
\draw (axis cs:4.5,7.5) node[
  scale=0.6,
  text=darkslategray38,
  rotate=0.0
]{0};
\draw (axis cs:5.5,7.5) node[
  scale=0.6,
  text=darkslategray38,
  rotate=0.0
]{7};
\draw (axis cs:6.5,7.5) node[
  scale=0.6,
  text=darkslategray38,
  rotate=0.0
]{1};
\draw (axis cs:7.5,7.5) node[
  scale=0.6,
  text=darkslategray38,
  rotate=0.0
]{189};
\draw (axis cs:8.5,7.5) node[
  scale=0.6,
  text=darkslategray38,
  rotate=0.0
]{32};
\draw (axis cs:0.5,8.5) node[
  scale=0.6,
  text=darkslategray38,
  rotate=0.0
]{39};
\draw (axis cs:1.5,8.5) node[
  scale=0.6,
  text=darkslategray38,
  rotate=0.0
]{2};
\draw (axis cs:2.5,8.5) node[
  scale=0.6,
  text=darkslategray38,
  rotate=0.0
]{51};
\draw (axis cs:3.5,8.5) node[
  scale=0.6,
  text=darkslategray38,
  rotate=0.0
]{4};
\draw (axis cs:4.5,8.5) node[
  scale=0.6,
  text=darkslategray38,
  rotate=0.0
]{7};
\draw (axis cs:5.5,8.5) node[
  scale=0.6,
  text=darkslategray38,
  rotate=0.0
]{2};
\draw (axis cs:6.5,8.5) node[
  scale=0.6,
  text=darkslategray38,
  rotate=0.0
]{0};
\draw (axis cs:7.5,8.5) node[
  scale=0.6,
  text=darkslategray38,
  rotate=0.0
]{25};
\draw (axis cs:8.5,8.5) node[
  scale=0.6,
  text=darkslategray38,
  rotate=0.0
]{713};
\end{axis}

\end{tikzpicture}
        \caption{Confusion matrix for \textit{gbert-base-comb context} trained on 20 shots on training set 3 with initial seed 123.}
        \label{fig:suppl:ablation_tests:confusiion-matrix-20shots-set2-seed123-gbert-base-comb-context}
    \end{figure}

            Hence, to test the ability of PET models to recognize, that the sample to classify is between the two \textit{[SEP]} token, we created nine artificial test samples by combining context paragraphs that are atypical for our dataset, as presented in Suppl. Fig. \ref{fig:suppl:ablation_tests:artificial-training-data}.\\
            If we use a \textit{gbert-base-comb context} model trained on 20 shots, the first sample in Suppl. Fig. \ref{fig:suppl:ablation_tests:artificial-training-data} is still incorrectly classified with $97\%$ as \textit{Anrede}. In contrast, the second sample is correctly classified with $99\%$ accuracy as \textit{Medikation}.\\
            Overall, 5/9 samples where still incorrectly classified as \textit{Anrede} class.\\
            We investigated another section class such as \textit{AllergienUnverträglichkeitenRisiken}, which typically only contains a single paragraph, too, we observed a similar behaviour. Often the context models incorrectly classify samples from the previous section \textit{Diagnosen} as \textit{AllergienUnverträglichkeitenRisiken}. E.g. \textit{ Z.n. Bandscheibenvorfall 11.09.1941 [SEP] - Z.n. Hodentorsion 11.09.1941 [SEP] Kardiovaskuläre Risikofaktoren: Arterielle Hypertonie, Hypercholesterinämie, positive Familienanamnese, Nikotinanamnese: nie}. These results raised the question: Are there often misclassifications at the first or final paragraph of a section class? The confusion matrix (Suppl. Fig. \ref{fig:suppl:ablation_tests:confusiion-matrix-20shots-set2-seed123-gbert-base-comb-context}) shows that typical false positives involve such patterns. For example:
            \begin{itemize}
                \item Diagnosen often misclassified as \textit{Anrede}
                \item \textit{Diagnosen} and \textit{Befunde} often misclassifed as \textit{AllergienUnverträglichkeitenRisiken}
                \item \textit{Medikation} and \textit{Zusammenfassung} often misclassifed as \textit{Abschluss}
            \end{itemize}
            This reveals, that contextualizing paragraphs can harm classification results for certain section classes. This is especially relevant for section classes, which usually contain single-paragraph samples. This suggests that in a few-shot learning scenario, smaller PLMs can have difficulty distinguishing testing instances from contexts, and hence do not sufficiently focus on the instances themselves.
        \subsubsection{Removing section titles from data}
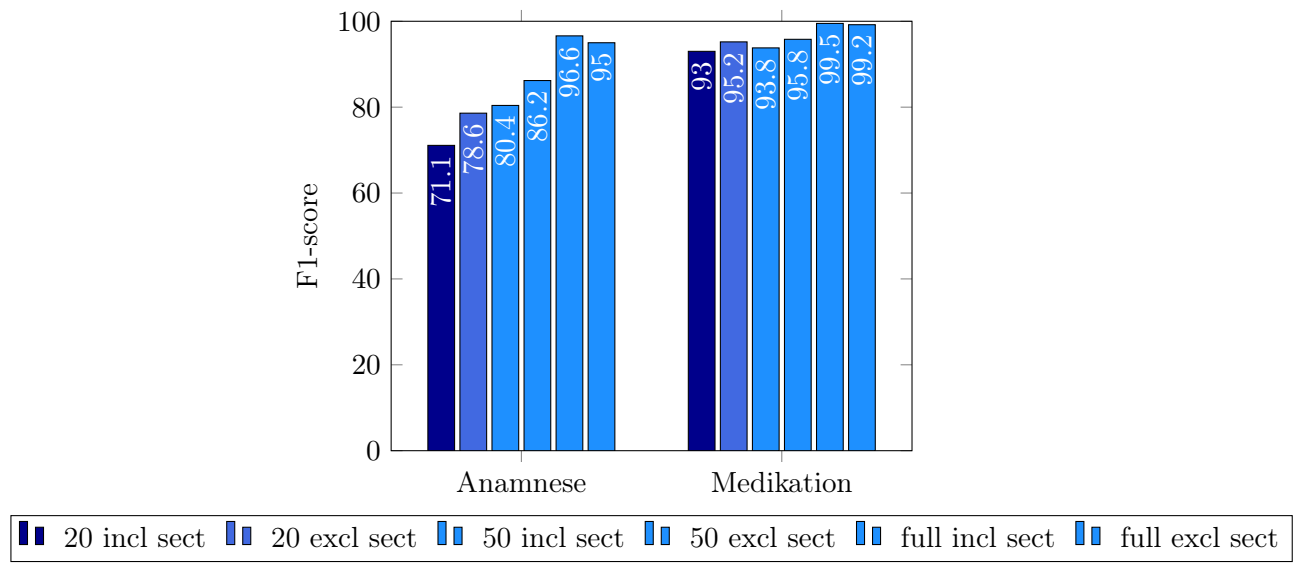
\begin{figure}
        \begin{tikzpicture}
            \begin{axis}[
                ybar,
                enlarge x limits=0.5,
                legend style={
                at={(0.5,-0.15)},
                anchor=north,
                legend columns=-1,
                column sep=0.5em,
                },
                ylabel={F1-score},
                symbolic x coords={Anamnese,Medikation},
                xtick=data,
                nodes near coords,
                nodes near coords align={vertical},
                ymax=100,
                ymin=0
                ]
                \addplot[fill=mydarkblue,nodes near coords style={rotate=90,left,color=white}] coordinates {(Anamnese,71.1) (Medikation,93)};
                \addplot[fill=myroyalblue,nodes near coords style={rotate=90,left,color=white}] coordinates {(Anamnese,78.6) (Medikation,95.2)};
                \addplot[fill=mydodgerblue,nodes near coords style={rotate=90,left,color=white}] coordinates {(Anamnese,80.4) (Medikation,93.8)};
                \addplot[fill=mydodgerblue,nodes near coords style={rotate=90,left,color=white}] coordinates {(Anamnese,86.2) (Medikation,95.8)};
                \addplot[fill=mydodgerblue,nodes near coords style={rotate=90,left,color=white}] coordinates {(Anamnese,96.6) (Medikation,99.5)};
                \addplot[fill=mydodgerblue,nodes near coords style={rotate=90,left,color=white}] coordinates {(Anamnese,95.0) (Medikation,99.2)};
                \legend{20 incl sect, 20 excl sect, 50 incl sect, 50 excl sect, full incl sect , full excl sect}
            \end{axis}
        \end{tikzpicture}            
        \caption{Comparing accuracy scores for \textit{gbert-large-comb context} including and excluding section titles. For reference we show results for SC model trained on full training sets for both scenarios.}
        \label{fig:suppl:ablation_tests:removed-section-titles-primary}
    \end{figure}

            We identified that our best performing model from the core experiments \textit{gbert-base-comb nocontext} using 20 shots frequently misclassified samples containing section titles of our primary classes. $32\%$ of the false negative samples of the \textit{Medikation} class contained either the text sequence \textit{Medikation bei Aufnahme:} or \textit{Medikation bei Entlassung:}. A similar classification error we observed for the \textit{Anamnese} class: $81\%$ of false negatives contain the text sequence \textit{Anamnese}. While in the training samples for  \textit{Medikation} we did not identify any section titles, there was a single title \textit{Anamnese:} in the training set of \textit{Anamnese}.\\
            Adding context and increasing model size could significantly avoid these kind of errors, still $5\%$ of the false negatives of the \textit{Medikation} class of our final model \textit{gbert-large-comb context} contained these kind of text sequences. Hence, we trained our final model \textit{gbert-large-comb context} on a modified training and test set, filtered by a list of the most common section titles (Suppl. Fig. \ref{fig:suppl:ablation_tests:filtered_section_titles}). In Suppl. Tab. \ref{tab:suppl:ablation_tests:large-context-remove-section-titles} shows, that accuracy could be increased over all few-shot sizes by approximately $2\%$. Suppl. Fig. \ref{fig:suppl:ablation_tests:removed-section-titles-primary} shows, that both primary classes can improve F1-scores for 20 and 50 shots. In contrast, the models trained on the full training set, slightly decrease in performance. This is not surprising, as in contrast to the few-shot sets, the full training set frequently contains section titles.\\
            However, it is important to note that these results can not be compared directly to the experimental results with included section titles, since we modified the training and test data set. Considering  experimental limitations in clinical routine, it may be beneficial to avoid the use of section titles as they can be often well identified through manual patterns and heuristics. This approach is especially relevant if only smaller PLMs are employed with strong sequence length restrictions due to limited resources.
        \subsubsection{Classifying \textit{nocontext} samples using a \textit{context} model}
\begin{figure}
        \centering
        \begin{subfigure}{0.9\textwidth}
            \centering
            \includegraphics[width=\textwidth]{images/shap/shap_gbert-comb-20-zusammenfassung-nocontext.png}
            \caption{}
            \label{fig:suppl:results:shap-gbert-large-base-comb-20-nocontext_subfigure_a}
        \end{subfigure}
        \begin{subfigure}{0.9\textwidth}
            \centering
            \includegraphics[width=\textwidth]{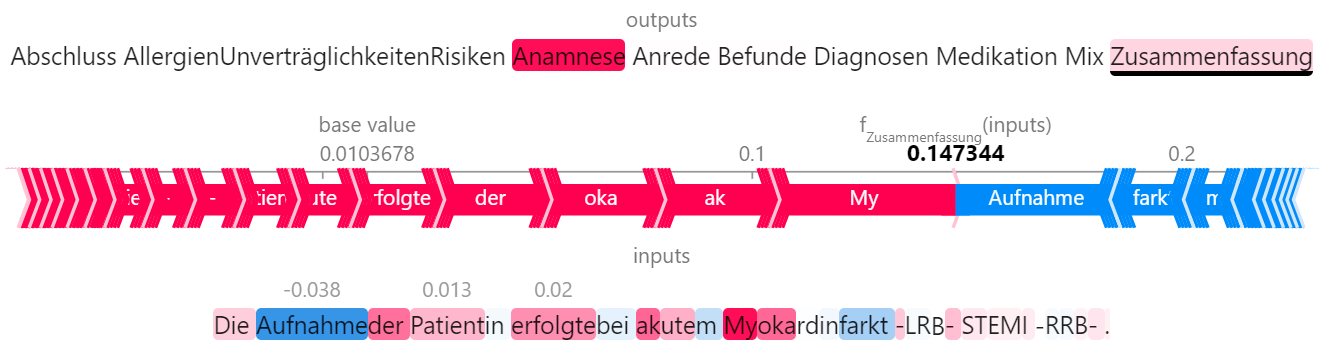}
            \caption{}
            \label{fig:suppl:results:shap-gbert-large-base-comb-20-nocontext_subfigure_b}
        \end{subfigure}
        \caption{Shapley values for predicted class \textit{Zusammenfassung} comparing (a) \textit{gbert-base-comb nocontext} and (b) \textit{gbert-large-comb nocontext} with 20 training shots. Shapley values with respect to predicted label (underlined). Shapley values per sub tokens.  \\Legend: \textcolor{red}{Red: positive contribution}, \textcolor{blue}{Blue: negative contribution}.}
        \label{fig:suppl:results:shap-gbert-large-base-comb-20-nocontext}
    \end{figure}

\begin{figure}
        \centering
        \begin{subfigure}{0.9\textwidth}
            \centering
            \includegraphics[width=\textwidth]{images/shap/shap_gbert-comb-20-zusammenfassung-context.png}
            \caption{}
            \label{fig:suppl:results:additional_experiments:combined_methods:shap-gbert-base-large-comb-20-context_subfigure_a}
        \end{subfigure}
        \begin{subfigure}{0.9\textwidth}
            \centering
            \includegraphics[width=\textwidth]{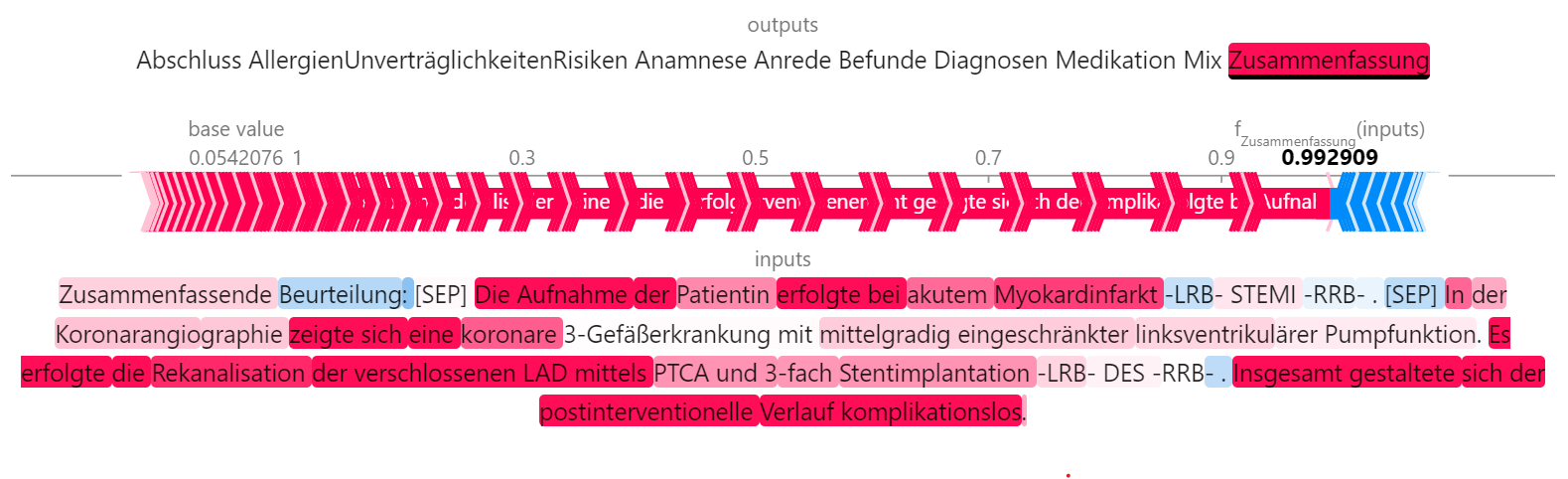}
            \caption{}
            \label{fig:suppl:results:additional_experiments:combined_methods:shap-gbert-base-large-comb-20-context_subfigure_b}
        \end{subfigure}
        \caption{Shapley values for gbert-large-comb, context for predicted class \textit{Zusammenfassung} comparing (a) \textit{gbert-base context} and (b) \textit{gbert-large context} with 20 training shots. Shapley values with respect to predicted label (underlined). Shapley values per sub tokens. \\Legend: \textcolor{red}{Red: positive contribution}, \textcolor{blue}{Blue: negative contribution}.}
        \label{fig:suppl:results:additional_experiments:combined_methods:shap-gbert-base-large-comb-20-context}
    \end{figure}

\begin{figure}
        \centering
        \begin{subfigure}{0.9\textwidth}
            \includegraphics[width=\textwidth]{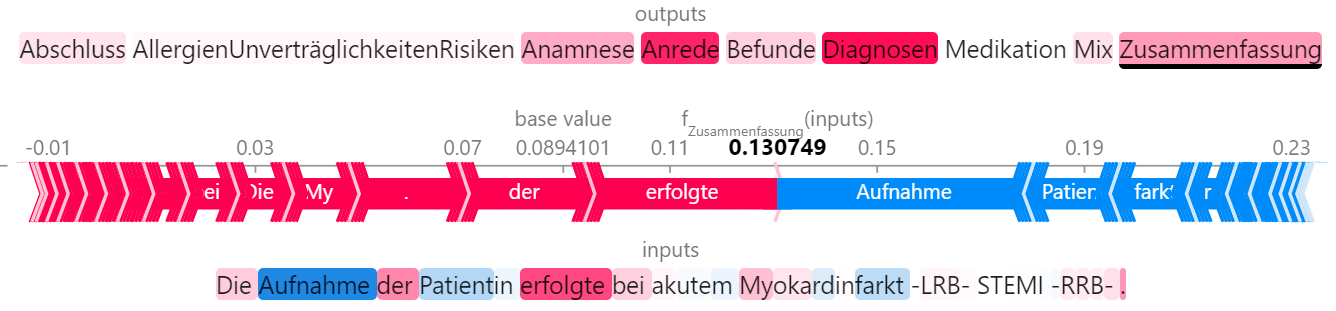}
    	    \caption{}
            \label{fig:suppl:ablation_tests:contextmodel_nocontextsample_subfigure_a}
        \end{subfigure}
        \begin{subfigure}{0.9\textwidth}
            \centering
            \includegraphics[width=\textwidth]{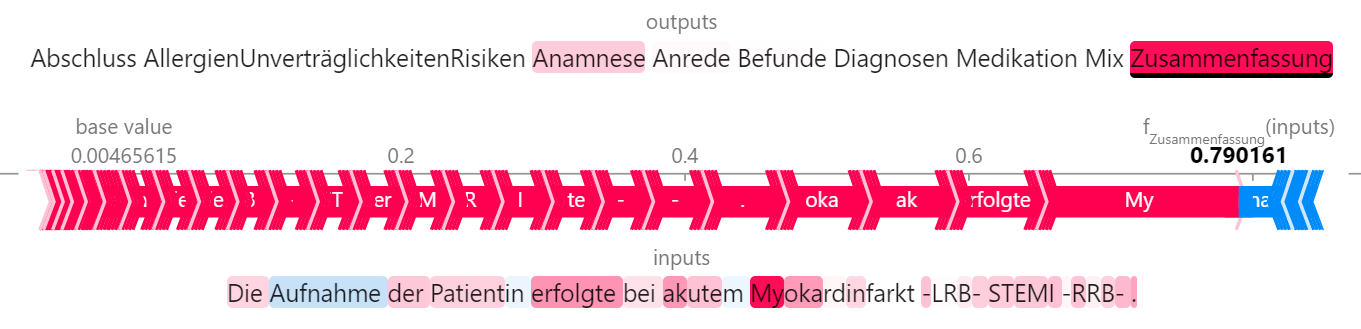}
            \caption{}
            \label{fig:suppl:ablation_tests:contextmodel_nocontextsample_subfigure_b}
        \end{subfigure}
        \caption{Comparing Shapley values for (a) \textit{gbert-base-comb} context model vs. (b) \textit{gbert-large-comb} context model using a sample without context. Shapley values with respect to predicted label (underlined). Shapley values per sub tokens. Legend: \textcolor{red}{Red: positive contribution}, \textcolor{blue}{Blue: negative contribution}.}
        \label{fig:suppl:ablation_tests:contextmodel_nocontextsample}
    \end{figure}

            In Suppl. Fig. \ref{fig:suppl:ablation_tests:contextmodel_nocontextsample} we show Shapley values of a sample without further context with the gold label \textit{Zusammenfassung} classified by (a) \textit{gbert-base-comb context} and (b) the \textit{gbert-large-comb context} model. \textit{Gbert-base-comb context} shows very similar token contributions with respect to \textit{Zusammenfassung} as \textit{gbert-base-comb nocontext} in \ref{fig:suppl:results:shap-gbert-large-base-comb-20-nocontext_subfigure_a}. But both base models incorrectly classify the sample.\\
            In contrast, \textit{gbert-large-comb context}, correctly assigns the \textit{Zusammenfassung} class with a probability of $79\%$. Adding context paragraphs increases this to $99\%$ (see \ref{fig:suppl:results:additional_experiments:combined_methods:shap-gbert-base-large-comb-20-context_subfigure_b}). Interestingly, most of the input token positively contribute to the correct class, with the exception of \textit{Aufnahme}. This is expected, as this token frequently negatively contributed to \textit{Zusammenfassung} in various experimental setups.
    \subsection{Hyperparameters}
        \label{section:suppl:hyperparameters}
        \textbf{Further pretraining}: We applied the following hyperparameters for pretraining experiments described in Section \ref{section:methods:pre-training}: vocabulary size: 30,000; maximum sequence length: 512;
        \begin{enumerate}
            \item task-adaptation:
                \begin{itemize}
                    \item  data: CARDIO:DE corpus
                    \item epochs: 100, batch size: 24, fp16: True, gradient accumulation steps: 4
                    \item 1×RTX6000 graphics processing unit (GPU) with 24 GB video random access memory (VRAM)
                    \item Training time: $\sim$ 2h
                \end{itemize}
             \item domain-adaptation:
                \begin{itemize}
                    \item data: 179,000 German doctor's letters $+$ GGPONC
                    \item epochs: 3, batch size: 16, fp16: True, gradient accumulation steps: 1
                    \item 2×RTX6000 GPUs with each 24 GB VRAM
                    \item Training time: $\sim$ 17h
                \end{itemize}
            \item combined: 
                \begin{itemize}
                    \item data: CARDIO:DE corpus
                    \item epochs: epochs: 100, batch size: 24, fp16: True, gradient accumulation steps: 4
                    \item 1×RTX6000 GPU with 24 GB VRAM
                    \item Training time: $\sim$ 2h
                \end{itemize}
        \end{enumerate}
        \textbf{PET and SC experiments}:
        \begin{itemize}
            \item All PET experiments were conducted on a single NVIDIA A40 GPU with $40$ GB VRAM. However, we also conducted PET experiments on NVIDIA P4 with $8$ GB VRAM using BERT-base models by only reducing evaluation batch size at inference time.
            \item Hyperparameters PET and SC: BERT-base models: training batch size $4$, evaluation batch size: $64$; BERT-large models: training batch size $4$, evaluation batch size $16$.
            \item Each experiment conducted with three different training sets and two random seeds (in total six setups).
            \item To increase comparability, we always selected models trained on training set $3$ and with random seed $123$ to investigate Shapley values.
        \end{itemize}

    \begin{figure}
        \centering
        \begin{tikzpicture}[node distance=0.6cm, auto]
            \node [draw, rectangle, minimum width=3cm, minimum height=0.8cm, align=center] (box1) {PLM};
            \node [draw, rectangle, right=of box1, minimum width=3cm, minimum height=0.8cm, align=center] (box2) {Pattern};
            \node [draw, rectangle, right=of box2, minimum width=3cm, minimum height=0.8cm, align=center] (box3) {Few-shot training data};
            \node [draw, rectangle, right=of box3, minimum width=3cm, minimum height=0.8cm, align=center] (box4) {Section class};
                
            \node [draw, rectangle, below=1.5cm of $(box1)!0.5!(box4)$, minimum width=5cm, minimum height=1.5cm, align=center, fill=mydarkblue, text=white] (box5) {Calculate most likely verbalizer tokens};
                
            \node [draw, rectangle, below=1.5cm of box5, minimum width=4.5cm, minimum height=0.6cm, align=center] (box6) {PLM vocabulary};
            \node [draw, rectangle, right=of box6, text width=4.5cm, align=center] (box6a) {
            Filtered by: non-words, words $\leq$ alphabetic chars,\\ 10,000 most frequent tokens in unlabeled training data
            };
                
            % Define arrows
            \draw [->] (box1) -- (box5) node[midway, right, xshift=2.5cm] {For each};
            \draw [->] (box2) -- (box5);
            \draw [->] (box3) -- (box5);
            \draw [->] (box4) -- (box5);
            \draw [<-] (box5) -- (box6)node[midway, right] {Given};
            \draw [<-] (box6) -- (box6a);
                
        \end{tikzpicture}
        \caption{The PETAL workflow: PETAL calculates the most likely verbalizer token per label for each (1) PLM, (2) prompt pattern, (3) few-shot training set. The verbalizer token must be part of the PLM's vocabulary.}
        \label{fig:suppl:methods:petal}
    \end{figure}
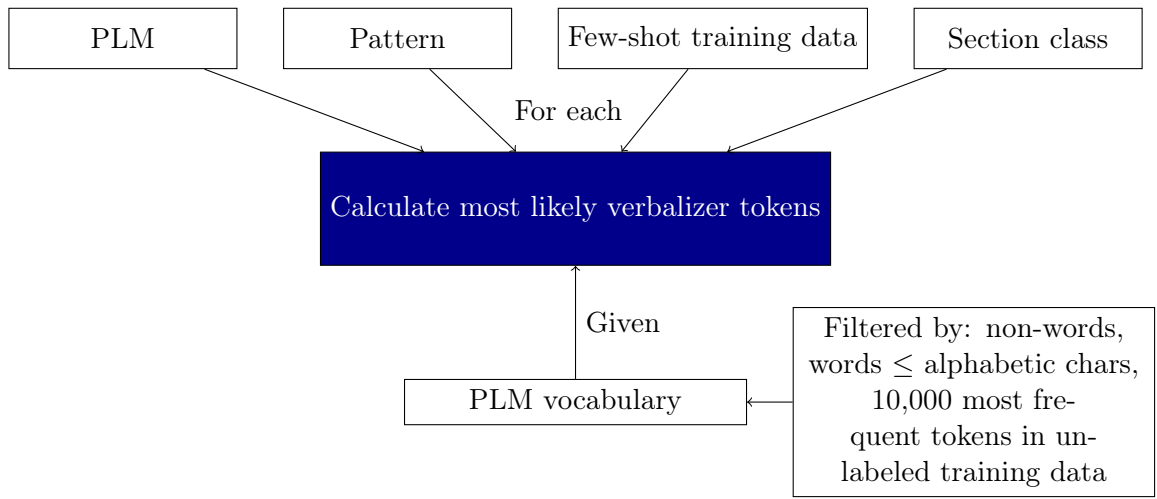

    \begin{figure}
        \centering
        \begin{verbatim}
            | - 10shots/
            |   | - set_1.csv
            |   | - set_2.csv
            |   | - set_3.csv
            |   | - unlabeled_1.csv
            |   | - unlabeled_2.csv
            |   | - unlabeled_3.csv
            | - holdout/
            |   | - full_holdout.csv
        \end{verbatim}
        \caption{Few-shot data: Example folder structure for the 10shot data set including the heldout data set.}
        \label{fig:suppl:experiments:fewshotfolder}
    \end{figure}

    \begin{figure}
        \centering
        \begin{tikzpicture}
            \begin{axis}[
                xlabel={Shot-size},
                ylabel={F1-score},
                xmin=0, xmax=6, % since you have 7 unique x values (including 'full')
                ymin=0, ymax=100,
                xtick={0,1,2,3,4,5,6},
                xticklabels={10,20,50,100,200,400,full}, 
                ytick={20,40,60,80,100},
                legend pos=south east,
                ymajorgrids=true,
                grid style=dashed,
            ]
            
            \addplot[
                color=blue,
                mark=square,
                ]
                coordinates {
                (0,50.3)(1,51.5)(2,67.3)(3,77.1)(4,78.6)(5,82.4)(6,88.1) % adjusted x coordinates
                };
                \addlegendentry{Anamnese}
            
            \addplot[
                color=red,
                mark=triangle,
                ]
                coordinates {
                (0,89.3)(1,89.9)(2,92.4)(3,95.4)(4,95.6)(5,97.5)(6,98.7) % adjusted x coordinates
                };
                \addlegendentry{Medikation}
            
            \end{axis}
            \end{tikzpicture}
        \caption{\textbf{Core experiments: Primary class $F1$-score for all shot sizes}. $F1$-score per few-shot sizes for primary classes with no context using \textit{gbert-base-comb nocontext}.}
        \label{fig:suppl:results:nocontext:primary_classes_allshots}
    \end{figure}
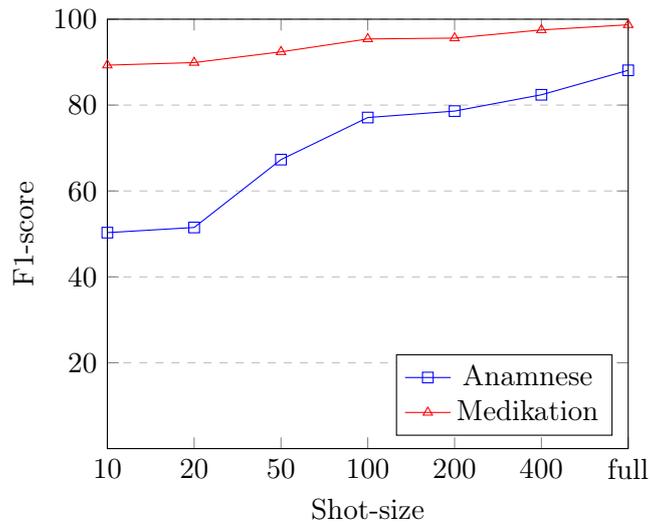

    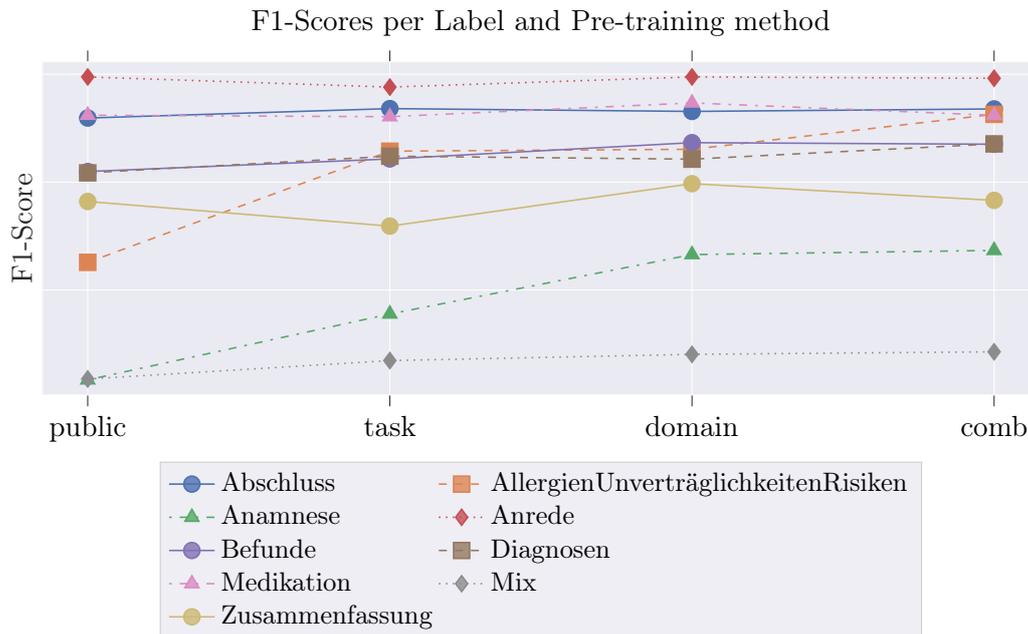
\begin{figure}
        \centering
        % This file was created with tikzplotlib v0.10.1.
\begin{tikzpicture}

\definecolor{darkkhaki204185116}{RGB}{204,185,116}
\definecolor{darkslategray38}{RGB}{38,38,38}
\definecolor{gray140}{RGB}{140,140,140}
\definecolor{gray14712096}{RGB}{147,120,96}
\definecolor{indianred1967882}{RGB}{196,78,82}
\definecolor{lavender234234242}{RGB}{234,234,242}
\definecolor{lightgray204}{RGB}{204,204,204}
\definecolor{lightslategray129114179}{RGB}{129,114,179}
\definecolor{mediumseagreen85168104}{RGB}{85,168,104}
\definecolor{orchid218139195}{RGB}{218,139,195}
\definecolor{peru22113282}{RGB}{221,132,82}
\definecolor{steelblue76114176}{RGB}{76,114,176}

\begin{axis}[
width=\textwidth, 
height=6cm,       
axis background/.style={fill=lavender234234242},
axis line style={white},
legend cell align={left},
legend style={
  fill opacity=0.8,
  draw opacity=1,
  text opacity=1,
  at={(0.5,-0.2)}, 
  anchor=north,    
  legend columns=2,
  draw=lightgray204,
  fill=lavender234234242,
  font=\small
},
tick align=outside,
title={F1-Scores per Label and Pre-training method},
x grid style={white},
xmajorgrids,
xmajorticks=true,
xmin=-0.15, xmax=3.15,
xtick style={color=darkslategray38},
xtick={0,1,2,3},
xtick={0,1,2,3},
xtick={0,1,2,3},
xtick={0,1,2,3},
xtick={0,1,2,3},
xtick={0,1,2,3},
xtick={0,1,2,3},
xtick={0,1,2,3},
xtick={0,1,2,3},
xticklabels={public, task, domain, comb}, 
y grid style={white},
ylabel=\textcolor{darkslategray38}{F1-Score},
ymajorgrids,
ymajorticks=false,
ymin=0.405279900660246, ymax=1.02305558740575,
ytick style={color=darkslategray38}
]
\addplot [semithick, steelblue76114176, mark=*, mark size=3, mark options={solid}]
table {%
0 0.918807042356625
1 0.9362850203874
2 0.931025361849684
3 0.935641383908988
};
\addlegendentry{Abschluss}
\addplot [semithick, peru22113282, dashed, mark=square*, mark size=3, mark options={solid}]
table {%
0 0.651280266052895
1 0.85733019453887
2 0.860875352297612
3 0.92566059388931
};
\addlegendentry{AllergienUnverträglichkeitenRisiken}
\addplot [semithick, mediumseagreen85168104, dash pattern=on 1pt off 3pt on 3pt off 3pt, mark=triangle*, mark size=3, mark options={solid}]
table {%
0 0.433360613694132
1 0.555527396097612
2 0.665679303538777
3 0.6733378454292
};
\addlegendentry{Anamnese}
\addplot [semithick, indianred1967882, dotted, mark=diamond*, mark size=3, mark options={solid}]
table {%
0 0.994974874371859
1 0.976119817415921
2 0.994974874371859
3 0.99248743718593
};
\addlegendentry{Anrede}
\addplot [semithick, lightslategray129114179, mark=*, mark size=3, mark options={solid}]
table {%
0 0.819623376570924
1 0.842974802758535
2 0.873131173355184
3 0.870228496637316
};
\addlegendentry{Befunde}
\addplot [semithick, gray14712096, dashed, mark=square*, mark size=3, mark options={solid}]
table {%
0 0.817130507069596
1 0.848045746091518
2 0.842489859319245
3 0.870810004925774
};
\addlegendentry{Diagnosen}
\addplot [semithick, orchid218139195, dash pattern=on 1pt off 3pt on 3pt off 3pt, mark=triangle*, mark size=3, mark options={solid}]
table {%
0 0.923976206701185
1 0.921249044778313
2 0.946819972089242
3 0.923894765365551
};
\addlegendentry{Medikation}
\addplot [semithick, gray140, dotted, mark=diamond*, mark size=3, mark options={solid}]
table {%
0 0.434953233272426
1 0.469146517259708
2 0.480698818849924
3 0.485407570863666
};
\addlegendentry{Mix}
\addplot [semithick, darkkhaki204185116, mark=*, mark size=3, mark options={solid}]
table {%
0 0.764083047511456
1 0.718422618869877
2 0.797079966801393
3 0.76619587124556
};
\addlegendentry{Zusammenfassung}
\end{axis}

\end{tikzpicture}
        \caption{F1-scores per label per pretraining method using \textit{gbert-base nocontext}}.
        \label{fig:suppl:discussion:f1-scores-per-label-per-pretraining}
    \end{figure}

    \begin{figure}
        \centering
        \includegraphics[width=\textwidth]{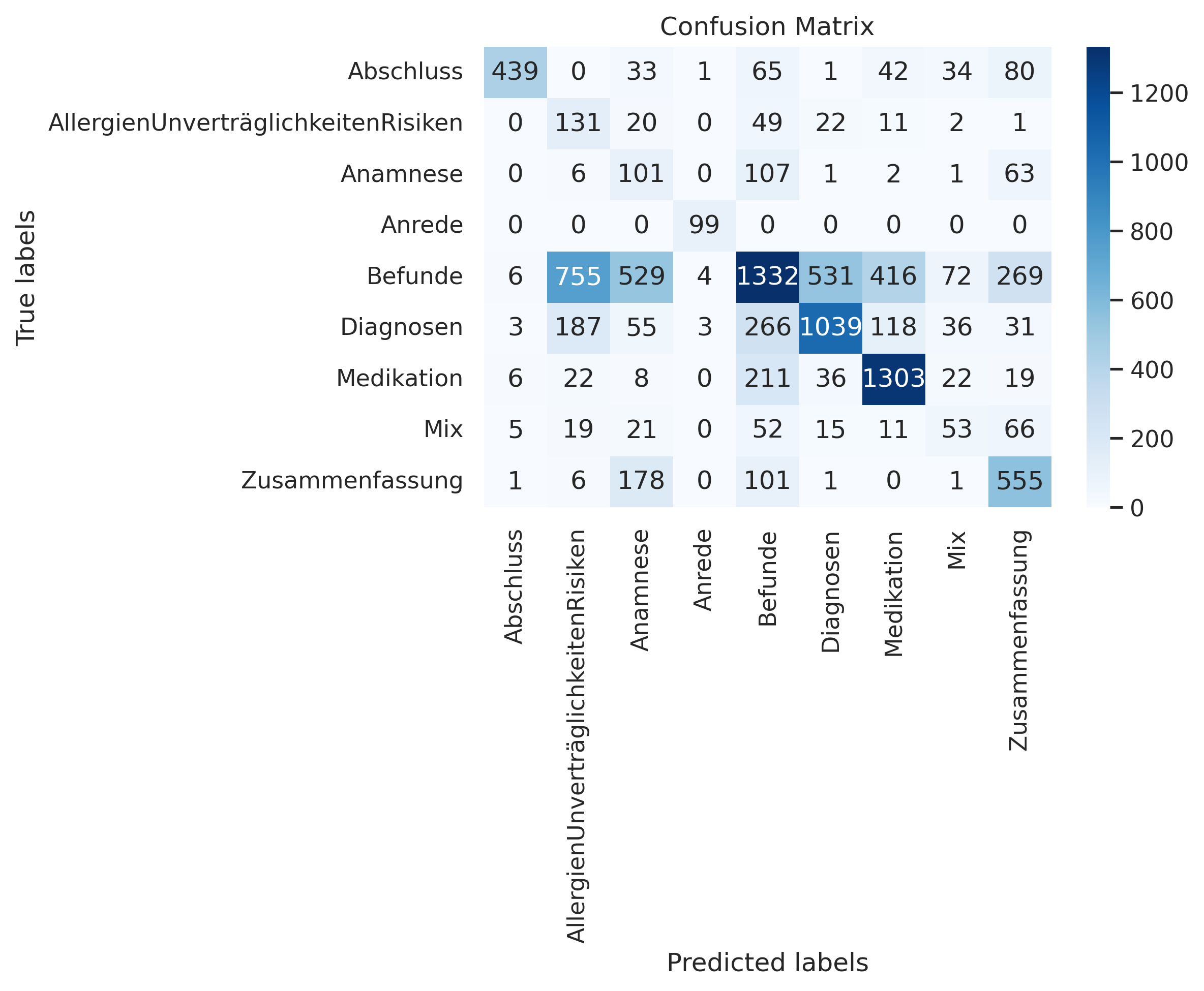}
        \caption{Confusion matrix for \textit{gbert-base} trained on 20 shots on training set 3 with initial seed 123.}
        \label{fig:supp:results:confusion_matrix_nocontext}
    \end{figure}

    \begin{figure}
        \centering
        \begin{subfigure}{0.9\textwidth}
            \centering
            \includegraphics[width=\textwidth]{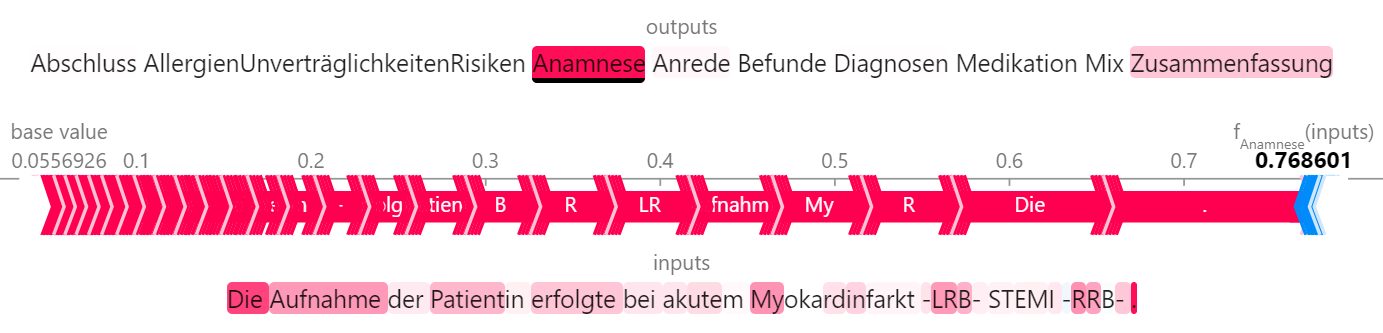}
            \caption{}
            \label{fig:suppl:results::shap-gbert-base-comb-20-nocontext_subfigure_a}
        \end{subfigure}
        \begin{subfigure}{0.9\textwidth}
            \centering
            \includegraphics[width=\textwidth]{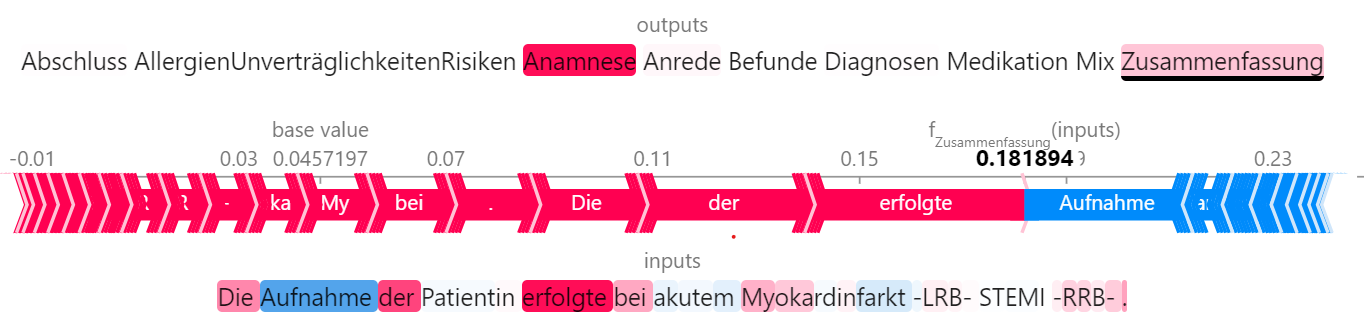}
            \caption{}
            \label{fig:suppl:results::shap-gbert-base-comb-20-nocontext_subfigure_b}
        \end{subfigure}
        \caption{Shapley values for \textit{gbert-base-comb nocontext} for predicted class comparing (a) \textit{Anamnese} and (b) \textit{Zusammenfassung} using 20 training shots. Shapley values with respect to predicted label (underlined). Shapley values per sub tokens. Legend: \textcolor{red}{Red: positive contribution}, \textcolor{blue}{Blue: negative contribution}.}
        \label{fig:suppl:results::shap-gbert-base-comb-20-nocontext}
    \end{figure}

    \begin{table}
        \centering
        \begin{tabular}{lcc}
            \toprule
            & Training set & Test set \\
            \midrule
            Anrede & 402 & 99 \\
            AktuellDiagnosen & 3,298 & 694 \\
            Diagnosen & 4,725 & 1,044 \\
            AllergienUnverträglichkeitenRisiken & 1,031 & 236 \\
            Anamnese & 1,188 & 281 \\
            AufnahmeMedikation & 2,058 & 593 \\
            KUBefunde & 4,194 & 1,105 \\
            Befunde & 9,636 & 2,519 \\
            EchoBefunde & 1,566 & 290 \\
            Labor & 55,420 & 12,220 \\
            Zusammenfassung & 3,645 & 843 \\
            Mix & 945 & 242 \\
            EntlassMedikation & 4,090 & 1,034 \\
            Abschluss & 2,805 & 695 \\
            \midrule
            Total & 95,003 & 21,895 \\
            \bottomrule
        \end{tabular}
        \caption{Number of samples per section class per CARDIO:DE corpus split.}
        \label{tab:suppl:data:count_sect}
    \end{table}

    \begin{table}
        \centering
        \begin{tabular}{llll}
            \toprule
            PLM & Pretrained & Method & Few-shots \\
            \midrule
            \multirow{4}{*}{gbert-base} & public & \multirow{8}{*}{PET\&SC} & \multirow{8}{*}{10, 20, 50, 100, 200, 400} \\
            & task & & \\
            & domain & & \\
            & comb & & \\
            \cmidrule{1-2}
            \multirow{4}{*}{medbertde-base} & public & & \\
            & task & & \\
            & domain & & \\
            & comb & & \\
            \bottomrule
        \end{tabular}
        \caption{Setup for core experiments: Experimental overview for our core experiments including PLMs, pretraining method, learning method and few-shot sizes.}
        \label{tab:suppl:methods:core_experiments}
    \end{table}

    \begin{table}
        \centering
        \begin{tabular}{lll}
            \toprule
            Shot size & All templates & Null prompt templates \\
            \midrule
            20 & 79.1 & 78.2 \\
            50 & 85.6 & 84.6 \\
            100 & 88.3 & 88.5 \\
            400 & 89.7 & 90 \\
            \midrule
            full (SC) & 96.7 & \\
            \bottomrule
        \end{tabular}
        \caption{\textbf{Null prompts:} accuracy scores for \textit{gbert-base-comb nocontext} PLMs using all templates or null prompts on four few-shot sizes.}
        \label{tab:suppl:results:additional_experiments:null_prompts}
    \end{table}

    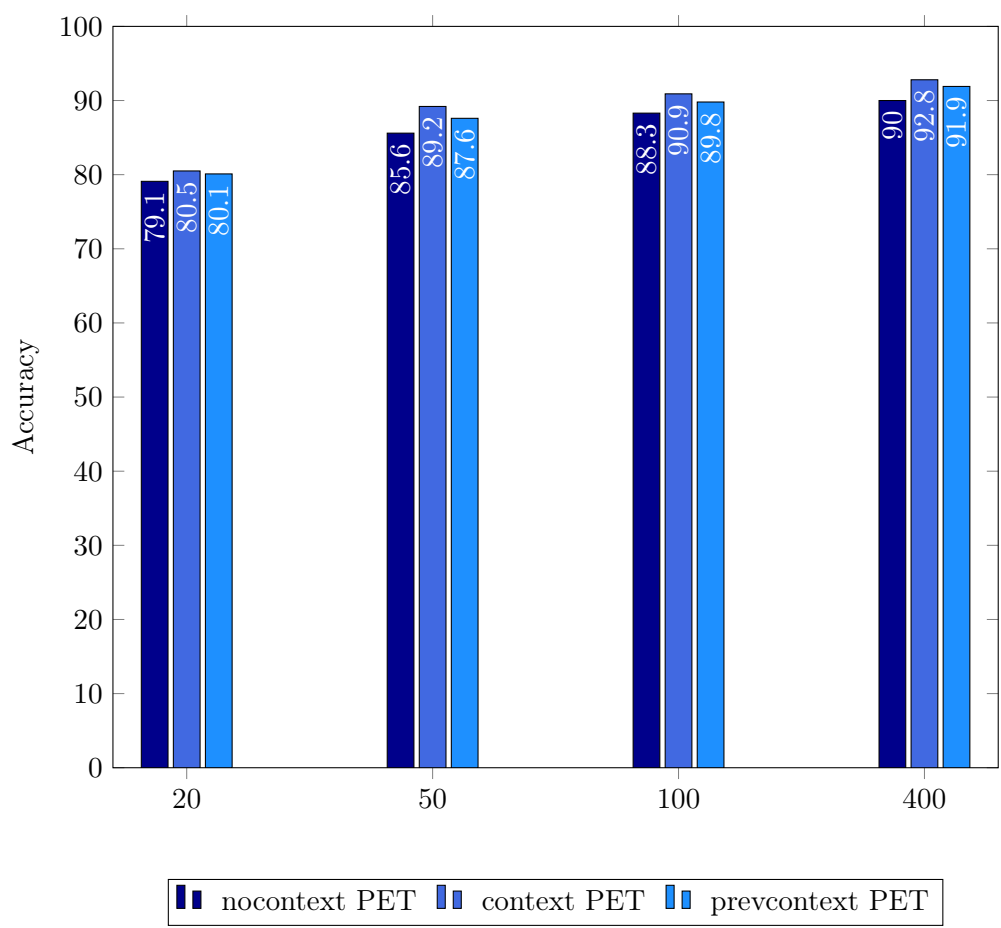
\begin{figure}
        \begin{tikzpicture}
            \begin{axis}[
                ybar,
                width=0.9\textwidth, % Adjust the value to increase or decrease the width
                legend style={
                at={(0.5,-0.15)},
                anchor=north,
                legend columns=-1,
                column sep=0.5em,
                },
                ylabel={Accuracy},
                symbolic x coords={20,50,100,400},
                xtick=data,
                nodes near coords,
                nodes near coords align={vertical},
                ymax=100, % Add this line to set the maximum value of the y-axis to 100
                ymin=0
                ]
                \addplot[fill=mydarkblue,nodes near coords style={rotate=90,left,color=white}] coordinates {(20,79.1) (50,85.6) (100,88.3) (400,90)};
                \addplot[fill=myroyalblue,nodes near coords style={rotate=90,left,color=white}] coordinates {(20,80.5) (50,89.2) (100,90.9) (400,92.8)};
                \addplot[fill=mydodgerblue,nodes near coords style={rotate=90,left,color=white}] coordinates {(20,80.1) (50,87.6) (100,89.8) (400,91.9)};
                \legend{nocontext PET, context PET, prevcontext PET}
            \end{axis}
        \end{tikzpicture}
        \caption{\textbf{Context types}: accuracy scores for different context types: (1) no context, (2) context, (3) prevcontext and few-shot sizes 20, 50, 100 and 400 using PET.}
        \label{fig:suppl:results:additional_experiments:context}
    \end{figure}

    \begin{table}
        \centering
        \begin{tabular}{lll}
            \toprule
            Shot size & gbert-large-comb & medbertde-base \\
            \midrule
            20 & 84.3 & 86.2 \\
            50 & 89.4 & 90.3 \\
            100 & 91.3 & 91.3 \\
            400 & 93.4 & 92.4 \\
            \bottomrule
        \end{tabular}
        \caption{Comparing \textit{gbert-large-comb context} and \textit{medbertde-base context} trained with all templates.}
        \label{tab:suppl:results:additional_experiments:medbertde}
    \end{table}

    \begin{figure}
        \centering
        \begin{subfigure}{0.9\textwidth}
            \centering
            \includegraphics[width=\textwidth]{images/shap/shap_gbert-comb-20-zusammenfassung-nocontext.png}
            \caption{}
            \label{fig:suppl:results:shap-gbert-base-comb-20-context_nocontext_subfigure_a}
        \end{subfigure}
        \begin{subfigure}{0.9\textwidth}
            \centering
            \includegraphics[width=\textwidth]{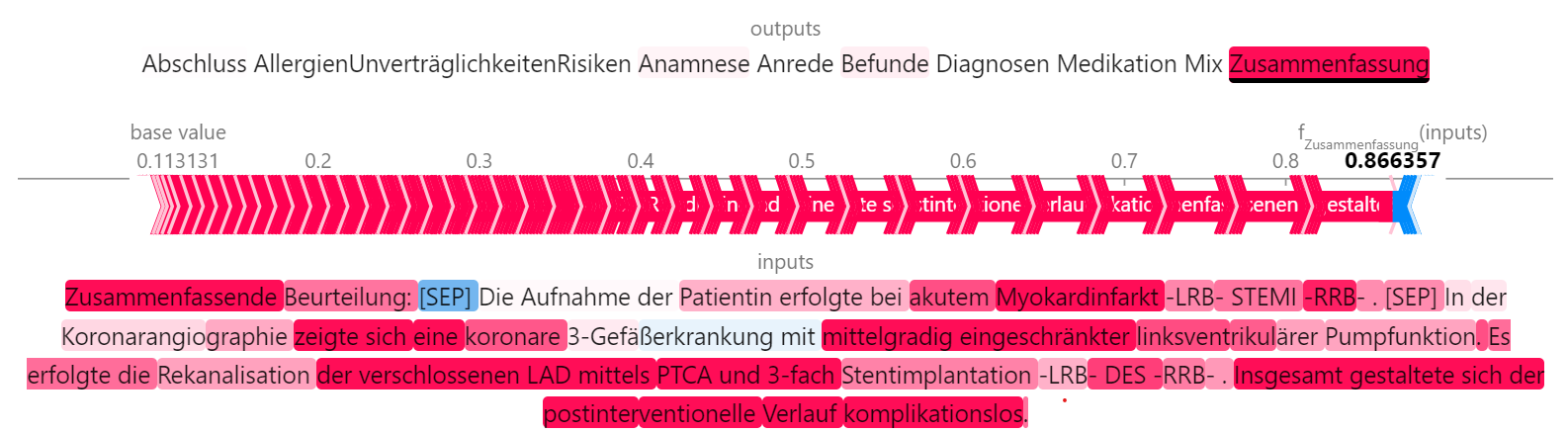}
            \caption{}
            \label{fig:suppl:results:shap-gbert-base-comb-20-context_nocontext_subfigure_b}
        \end{subfigure}
        \caption{Shapley values for \textit{gbert-base-comb context} for predicted class Zusammenfassung comparing (a) \textit{gbert-base nocontext} and (b) \textit{gbert-base context} with 20 training shots. Shapley values with respect to predicted label (underlined). Shapley values per sub tokens. \\Legend: \textcolor{red}{Red: positive contribution}, \textcolor{blue}{Blue: negative contribution}.}
        \label{fig:suppl:results:shap-gbert-base-comb-20-context_nocontext}
    \end{figure}

    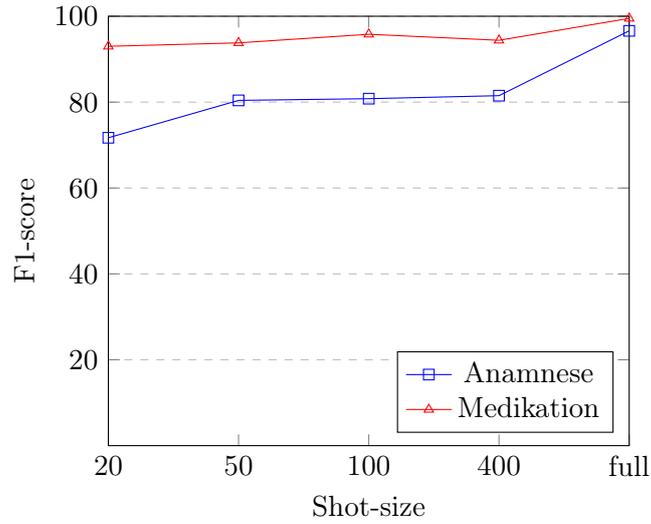
\begin{figure}
        \centering
        \begin{tikzpicture}
            \begin{axis}[
                xlabel={Shot-size},
                ylabel={F1-score},
                xmin=0, xmax=4,
                ymin=0, ymax=100,
                xtick={0,1,2,3,4},
                xticklabels={20,50,100,400,full}, 
                ytick={20,40,60,80,100},
                legend pos=south east,
                ymajorgrids=true,
                grid style=dashed,
            ]
            
            \addplot[
                color=blue,
                mark=square,
                ]
                coordinates {
                (0,71.7)(1,80.4)(2,80.8)(3,81.5)(4,96.6) 
                };
                \addlegendentry{Anamnese}
            
            \addplot[
                color=red,
                mark=triangle,
                ]
                coordinates {
                (0,93)(1,93.8)(2,95.8)(3,94.4)(4,99.5)
                };
                \addlegendentry{Medikation}
            
            \end{axis}
            \end{tikzpicture}
        \caption{\textbf{Additional experiments: Primary class $F1$-score for all shot sizes}: Accuracy scores per few-shot sizes for primary classes using \textit{gbert-large-comb context}.}
        \label{fig:suppl:results:context:primary_classes_allshots}
    \end{figure}

    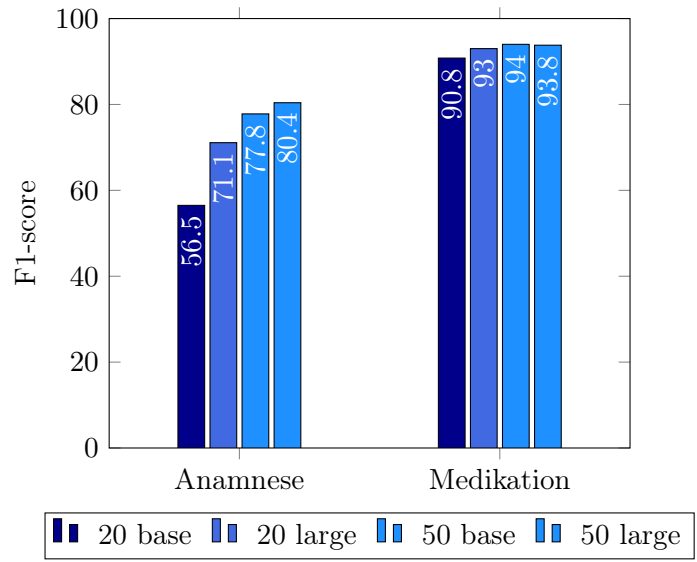
\begin{figure}
        \centering
        \begin{tikzpicture}
            \begin{axis}[
                ybar,
                enlarge x limits=0.5,
                legend style={
                at={(0.5,-0.15)},
                anchor=north,
                legend columns=-1,
                column sep=0.5em,
                },
                ylabel={F1-score},
                symbolic x coords={Anamnese,Medikation},
                xtick=data,
                nodes near coords,
                nodes near coords align={vertical},
                ymax=100,
                ymin=0
                ]
                \addplot[fill=mydarkblue,nodes near coords style={rotate=90,left,color=white}] coordinates {(Anamnese,56.5) (Medikation,90.8)};
                \addplot[fill=myroyalblue,nodes near coords style={rotate=90,left,color=white}] coordinates {(Anamnese,71.1) (Medikation,93.0)};
                \addplot[fill=mydodgerblue,nodes near coords style={rotate=90,left,color=white}] coordinates {(Anamnese,77.8) (Medikation,94.0)};
                \addplot[fill=mydodgerblue,nodes near coords style={rotate=90,left,color=white}] coordinates {(Anamnese,80.4) (Medikation,93.8)};
                \legend{20 base, 20 large, 50 base, 50 large}
            \end{axis}
        \end{tikzpicture}
        \caption{Combining best performing methods: comparing accuracy scores for \textit{gbert-large-comb context} vs. \textit{gbert-base-comb context} with all templates on two few-shot sizes for primary classes.}
        \label{fig:suppl:results:additional_experiments:combined_methods:primary_classes:nocontext_vs_context}
    \end{figure}

    \begin{figure}
        \centering
        \begin{tikzpicture}
            \begin{axis}[
                ybar,
                enlarge x limits=0.5,
                legend style={
                at={(0.5,-0.15)},
                anchor=north,
                legend columns=-1,
                column sep=0.5em,
                },
                ylabel={F1-score},
                symbolic x coords={Anamnese,Medikation},
                xtick=data,
                nodes near coords,
                nodes near coords align={vertical},
                ymax=100,
                ymin=0
                ]
                \addplot[fill=mydarkblue,nodes near coords style={rotate=90,left,color=white}] coordinates {(Anamnese,71.1) (Medikation,93.0)};
                \addplot[fill=myroyalblue,nodes near coords style={rotate=90,left,color=white}] coordinates {(Anamnese,65.9) (Medikation,94.8)};
                \addplot[fill=mydodgerblue,nodes near coords style={rotate=90,left,color=white}] coordinates {(Anamnese,80.4) (Medikation,93.8)};
                \addplot[fill=mydodgerblue,nodes near coords style={rotate=90,left,color=white}] coordinates {(Anamnese,77.2) (Medikation,95.4)};
                \addplot[fill=mydodgerblue,nodes near coords style={rotate=90,left,color=white}] coordinates {(Anamnese,96.6) (Medikation,99.5)};
                \legend{20 gbert, 20 medbertde, 50 gbert, 50 medbertde, full gbert}
            \end{axis}
        \end{tikzpicture}
        \caption{Comparing \textit{gbert-large-comb context} and \textit{medbertde-base context} trained with all templates. F1-score per primary label.}
        \label{fig:suppl:results:additional_experiments:combined_methods:primary_classes:medbertde}
    \end{figure}
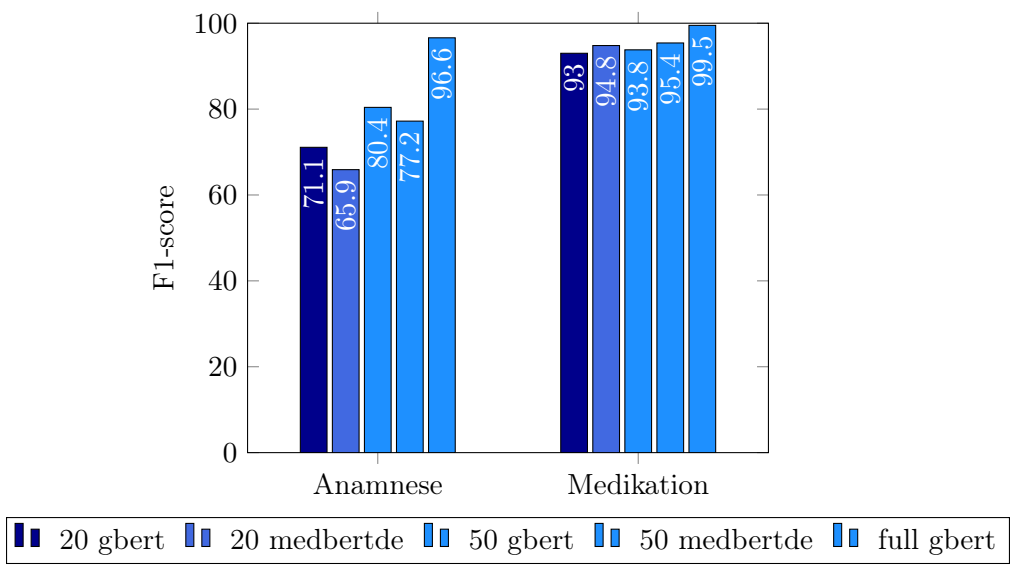

% Ablation tests

    \begin{figure}
        \centering
        \begin{myfigure}{}
            [SEP] über Ihre Patientin Frau Martina Mustermann geboren am 12.12.1999 wh. 2000 Musterstadt Musterstr. 1 die sich am in unserer Ambulanz vorstellte. [SEP] \textbf{Anamnese:} [SEP] Die Vorstellung erfolgte als Überweisung durch den niedergelassenen Kardiogen bei ED Vorhofflimmern.\\

            [SEP] über Ihre Patientin HErr Max Mustermann geboren am 12.12.1999 wh. 2000 Musterstadt Musterstr. 1 die sich am in unserer Ambulanz vorstellte. [SEP] \textbf{Medikation:} [SEP] Carvedilol 3,125 mg 1-0-1
        \end{myfigure}
        \caption{Two artificial training samples with atypical co-occuring context paragraphs. In the first sample, the section title \textit{Anamnese} follows immediately after a \textit{Anrede} sample. In the second example, \textit{Medikation} follows after a \textit{Anrede} sample. Usually \textit{Anrede} is followed by \textit{Diagnose}, rarely by \textit{Anamnese} and never in our data set by \textit{Medikation}.}
        \label{fig:suppl:ablation_tests:artificial-training-data}
    \end{figure}

    \begin{figure}
        \centering
        \begin{myfigure}{}
                Anamnese:, Diagnosen:, Zusammenfassung:, Körperlicher Untersuchungsbefund:, Aktuell:, Labor:, Ruhe-EKG:, Procedere:, Medikation bei Aufnahme:, Therapieempfehlung:, Beurteilung:, Therapieempfehlung -LRB- von kardiologischer Seite -RRB- :, Befund und Beurteilung:, Transthorakale Echokardiographie:, Maßnahmen:, Befund:, Aktuelle Medikation:, Beurteilender Abschnitt:, Beschreibender Abschnitt:, Zusammenfassende Beurteilung:, Belastungs-EKG:, Kultureller Befund:, Medikation bei Entlassung:, Lokalbefund:, Körperliche Untersuchung:, Allergien:, Ruhe-EKG bei Aufnahme:, Oral:, Kardiovaskuläre Risikofaktoren:, Nächster Termin/Kontrolle:, Procedere/Termine:, Aktuelle Therapie:, Diagnose:, Echokardiographie:, Bisherige Medikation:, Farbduplexsonographie der Gefäße der rechten Leiste:, Kapilläre Blutgasanalyse:, Sonstige Diagnosen:, Therapieempfehlung von kardiologischer Seite:, Nächster Termin/Prozedere:, Befund/Zusammenfassung:, Kommentar:, Indikation für stationären Herzkatheter:, EKG:, Spirometrie:, Wichtig:, Medikation:, Langzeit-EKG vom B-DATE:, Aktuelle/Bisherige Medikation:, Befund / Zusammenfassung:
        \end{myfigure}
        \caption{List of most common section titles: We generated this list by filtering the data set by short sequences including a single ":" at the end.}
        \label{fig:suppl:ablation_tests:filtered_section_titles}
    \end{figure}

    \begin{table}
        \centering
        \begin{tabular}{lllll}
            \toprule
            Shot size & PET including section titles & PET excl. section titles \\
            \midrule
            20 & 84.3 & 86.7 \\
            50 & 89.4 & 91.1 \\
            100 & 91.3 & 93.4 \\
            400 & 93.4 & 95.8 \\
            \bottomrule
        \end{tabular}
        \caption{F1-score results using \textit{gbert-large-comb context} trained with and without section titles in training and test data.}
        \label{tab:suppl:ablation_tests:large-context-remove-section-titles}
    \end{table}

\end{document}